\documentclass{article}

\usepackage{microtype}
\usepackage{graphicx}
\usepackage{subfigure}
\usepackage{booktabs} 

\usepackage{hyperref}

\usepackage[accepted]{icml2024}

\usepackage{amsmath}
\usepackage{amssymb}
\usepackage{mathtools}
\usepackage{amsthm}

\usepackage[capitalize,noabbrev]{cleveref}

\theoremstyle{plain}
\newtheorem{theorem}{Theorem}[section]

\newtheorem{lemma}[theorem]{Lemma}

\theoremstyle{definition}
\newtheorem{definition}[theorem]{Definition}

\theoremstyle{remark}

\usepackage[textsize=tiny]{todonotes}

\icmltitlerunning{Boosting Offline Optimizers with Surrogate Sensitivity}

\usepackage{natbib}  
\usepackage{caption} 

\usepackage{amssymb}
\usepackage{amsmath}
\usepackage{amsfonts}

\usepackage{graphicx}
\usepackage{subcaption}
\usepackage{makecell}

\usepackage{booktabs} 
\usepackage{multirow}

\DeclareMathOperator*{\argmax}{arg\,max}
\DeclareMathOperator*{\argmin}{arg\,min}

\newcommand{\ourmethod}{\texttt{BOSS}}
\newcommand{\variantmethod}{\texttt{BOSS-2}}
\newcommand{\cuong}[1]{\textcolor{black}{#1}}

\begin{document}

\twocolumn[
\icmltitle{Boosting Offline Optimizers with Surrogate Sensitivity}



\icmlsetsymbol{equal}{*}

\begin{icmlauthorlist}
\icmlauthor{Manh Cuong Dao}{hust}
\icmlauthor{Phi Le Nguyen}{hust}
\icmlauthor{Thao Nguyen Truong}{aist}
\icmlauthor{Trong Nghia Hoang}{wsu}
\end{icmlauthorlist}

\icmlaffiliation{hust}{School of Information and Communications Technology, Hanoi University of Science and Technology, Hanoi, Vietnam}
\icmlaffiliation{aist}{National Institute of Advanced Industrial Science and Technology, Tokyo, Japan}
\icmlaffiliation{wsu}{School of Electrical Engineering and Computer Science, Washington State University, Washington, USA}

\icmlcorrespondingauthor{Trong Nghia Hoang}{trongnghia.hoang@wsu.edu}

\icmlkeywords{Offline Optimization, Machine Learning}

\vskip 0.3in
]



\printAffiliationsAndNotice{}  

\begin{abstract}
Offline optimization is an important task in numerous material engineering domains where online experimentation to collect data is too expensive and needs to be replaced by an in silico maximization of a surrogate of the black-box function. Although such a surrogate can be learned from offline data, its prediction might not be reliable outside the offline data regime, which happens when the surrogate has narrow prediction margin and is (therefore) sensitive to small perturbations of its parameterization. This raises the following questions: (1) how to regulate the sensitivity of a surrogate model; and (2) whether conditioning an offline optimizer with such less sensitive surrogate will lead to better optimization performance. To address these questions, we develop an optimizable sensitivity measurement for the surrogate model, which then inspires a sensitivity-informed regularizer that is applicable to a wide range of offline optimizers. This development is both orthogonal and synergistic to prior research on offline optimization, which is demonstrated in our extensive experiment benchmark.
\end{abstract}

\section{Introduction}
\label{sec:intro}
Finding material designs that maximize a set of desirable properties is a fundamental task in material engineering. Historically, these design problems were frequently tackled through online experimentation, which can be exceedingly labor-intensive, time-consuming, and often impractical. To avoid such expenses, offline optimization \citep{BrookeICML19,TrabuccoICML21,trabucco2022design, nguyen2023expt, krishnamoorthy2022generative} has emerged as a computational alternative that leverages past experiment results to predict properties of unseen material candidates without running actual experiments. This is achieved via (1) fitting a parameterized model on such past data relating the material input with its output properties; and (2) finding an input optimizer with respect to the learned parameterization.

Naively, such \emph{in silico} approach would trivialize the optimal design problem into a vanilla application of gradient ascent and supervised learning. However, in practice, the prediction of such vanilla surrogate might not be reliable outside the offline data regime \citep{ClaraNeurIPs20}. Often, its prediction can become highly erratic at out-of-distribution data regimes, misguiding the optimization process toward sub-optimal candidates. This happens when the surrogate has narrow prediction margin at those out-of-distribution input regimes where a small perturbation to the model weights might cause a significant change in its prediction. This is potentially due to the fact that without relevant data, the training process does not have any mechanisms to recognize and avoid moving toward such sensitive model candidates.

To date, while most existing approaches addressing this problem have proposed numerous strategies to avoid making such spurious predictions during extrapolation (due to high model sensitivity), their conditioning techniques were not built on a direct characterization of sensitivity.  For example, \citet{FuICLR21,KumarNeurIPs20,TrabuccoICML21,trabucco2022design} and \citet{YuNeurIPs21} reduce their surrogate estimations at out-of-distribution inputs, while \citet{BrookeICML19} and \citet{ClaraNeurIPs20} condition on domain-specific properties under which sampled inputs will have high performance. Intuitively, model sensitivity could be reduced where the conditioning happens, which is however either surrogate- or search-specific. As a result, the conditioning is localized depending on how these out-of-distribution inputs and domain-specific properties are specified. It therefore remains unclear what impact such localized conditioning has on the overall model sensitivity. More importantly, it is also unclear whether in such approaches, the surrogate is well-conditioned at the regions where the search might visit. Otherwise, it would be less effective if the surrogate is only well-conditioned in regions where the search rarely visits. In this view, the core issue here is the lack of a model-agnostic characterization of sensitivity, which does not depend on the specifics of either the search or surrogate models. Such characterization could play the role of a negotiating medium that coordinates the conditioning between the search and surrogate models.

This prompts two fundamental inquiries: (1) how to explicitly characterize and regulate the sensitivity of a surrogate model; and (2) whether such sensitivity-regulated approach can be agnostic of the search and surrogate model specifications, allowing it to be seamlessly integrated into both the search process and surrogate models to enhance the overall optimization performance. To shed light on these matters, our paper formalizes a model-agnostic sensitivity measure which can be optimized to coordinate the surrogate and search biases so that the surrogate is conditioned to be risk-averse wherever the search goes. This is substantiated with the following technical contributions:

{\bf 1.} We develop a novel concept of model sensitivity in terms of how often its output would change beyond a user-specified threshold under small model perturbations. Intuitively, if a model's output is sensitive to such perturbations, its prediction margin must be narrow and have high variance, which means the chance that it would fit well the unseen part of the oracle function is low. The developed sensitivity notion therefore provides a quantifiable risk assessment that can be exploited by both the surrogate and search models (Section~\ref{sec:method}).

{\bf 2.} We show that our model sensitivity measurement is optimizable and can be used as a regularizer for a diverse range of existing offline optimizers which either define their search based on derivatives of their surrogate models or couple both search and surrogate models in a single differentiable loss function. To enable this, we also develop a numerical algorithm to tractably and effectively optimize our sensitivity-informed regularizers (Section~\ref{sec:algorithm}). The pseudocode of our \underline{b}oosting framework for offline \underline{o}ptimizers via \underline{s}urrogate \underline{s}ensitivity (\ourmethod) is detailed in Algorithm~\ref{alg:pseudo-code}.

{\bf 3.} We demonstrate the empirical efficiency of the proposed regularization method on a wide variety of benchmark problems, which shows consistently that its synergistic performance boost on existing offline optimizers is significant. Overall, our results corroborate our earlier hypothesis that a model-agnostic characterization of sensitivity can help coordinate the conditioning between the search and surrogate models better, which is evident from the consistent performance improvement across many baselines and optimization tasks (Section~\ref{sec:experiment}).

{\bf 4.} For clarity, we also provide a concise review of the existing literature in Section~\ref{sec:related}.

\section{Related Work}
\label{sec:related}
In numerous material engineering fields, such as molecular structure, robot morphology, and protein design, the primary objective is to discover the optimal design that maximizes performance based on specific criteria. A significant challenge in addressing this problem arises from the fact that the relationship between a design and its performance is a black-box function, which requires expensive experiments or simulations to evaluate the performance output of each candidate input. Consequently, finding the optimal design is equivalent to optimizing a black-box function whose derivative information is not accessible.

Furthermore, sampling data from this black-box function also requires excessive laboring cost of conducting biophysical experiments, which makes existing derivative-free methods, such as random gradient estimation \citep{WangAISTATS18} or Bayesian optimization \citep{snoek2012practical}, not economically viable as they often require sampling a large amount of data. This has inspired a new paradigm of offline optimization which learns an explicit model parameterization that explains well the relationship between input candidates and their corresponding experiment results in a past dataset. 

\begin{figure*}[bt]
    \centering
    \includegraphics[width=0.7\linewidth]{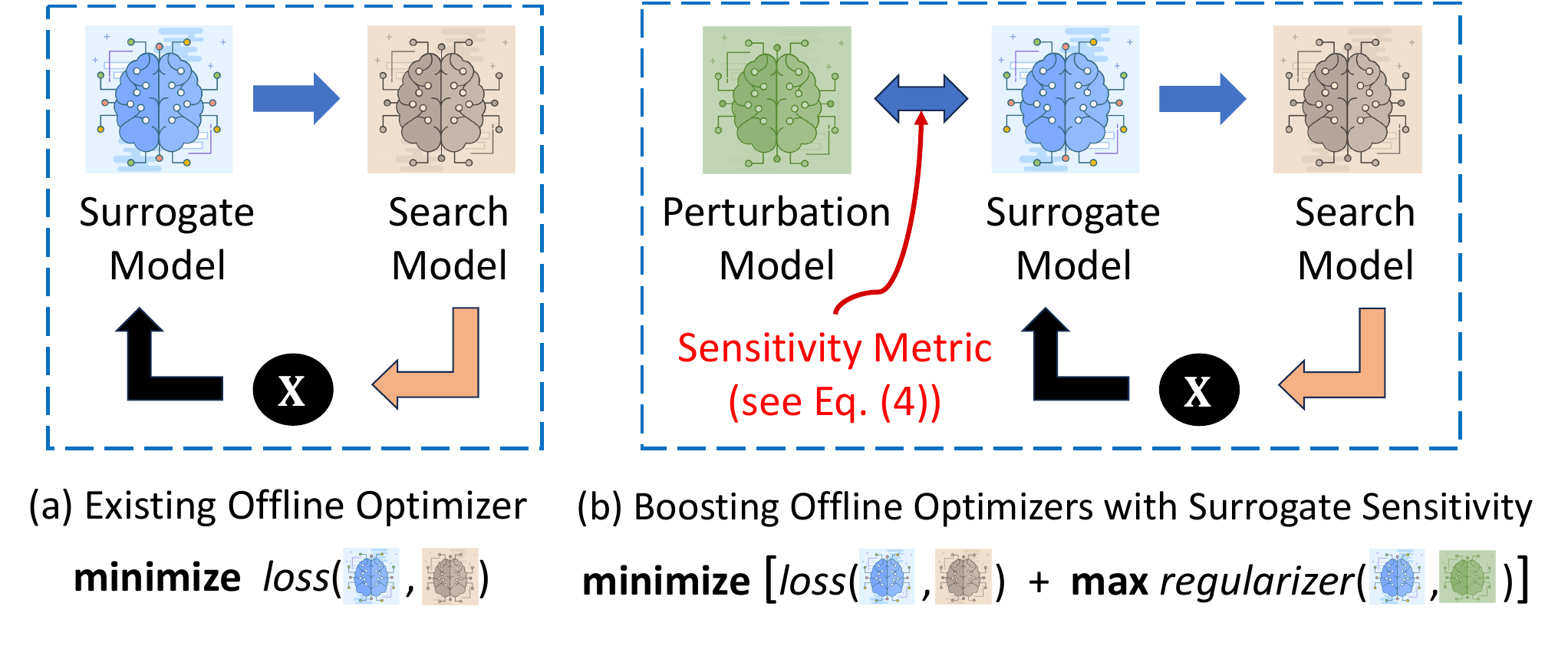}
    \caption{Workflows of (a) existing offline optimizers; and (b) our sensitivity-informed regularized optimizer (\ourmethod) which regulates the training workflow of existing offline optimizers with a new sensitivity metric -- see Definition~\ref{def:1}. Our regularizer is generic and can be applied to most existing offline optimizer workflows to boost their performance as demonstrated in Section~\ref{sec:experiment}.}
    \label{fig:method_overview}
    
\end{figure*}

To date, there have been several solutions proposed for offline optimization, which mostly focus on encoding some conservative preferences in either the search process or the (surrogate) training process. Such conservative preferences often aim to compensate for potential erratic function approximation so as to avoid false optimism during extrapolation. 
\citet{TrabuccoICML21} forces the model to underestimate the output value of input candidates found during early iterations of gradient updates (deemed out-of-distribution), whereas \citet{FuICLR21} maximizes the normalized data likelihood to reduce uncertainty in 
prediction. \citet{MinhICML24} matches the gradient fields of the oracle and surrogate, and shows that the optimization performance is directly bounded by the gradient gap. \citet{YuNeurIPs21} adopts techniques in model pre-training and adaptation to enforce a criteria of local smoothness. 

Alternatively, \citet{BrookeICML19,ClaraNeurIPs20} and \citet{YassineAAAI24} focus instead on conditioning the search process. Under this paradigm, the search model is represented as a distribution conditioned on rare event of achieving high oracle performance or an adaptive gradient update policy with learnable parameters~\cite{YassineAAAI24}, which is substantiated using different approaches. For instance, \citet{BrookeICML19} models such conditioned distribution via an adversarial zero-sum game while \citet{KumarNeurIPs20}~learns an inverse mapping from the performance output to the input design using conditional generative adversarial network \citep{Mehdi14}, from which design candidates performing at least as good as the example candidates in the offline dataset can be sampled. However, as mentioned previously, the implicit conditioning of these approaches is either search- or surrogate-specific, which does not coordinate well between the search and (surrogate) training processes. As such, the risk of following a search model will depend on how accurate the conditioning is at out-of-distribution inputs. This depends on the specific of the algorithm that was adopted, which has neither been defined nor investigated.


\section{Sensitivity-Guided Offline Optimization}
\label{sec:method}
In what follows, we will define the problem setting for offline optimization and introduce key notations (Section~\ref{sec:notation}). We formalize the concept of model sensitivity (Section~\ref{sec:trustability}) and develop a sensitivity-informed regularizer for existing offline optimizers (Section~\ref{sec:regularizer}). For clarity, an overview of our solution workflow is visualized in Figure ~\ref{fig:method_overview}.


\subsection{Problem Setting and Notations}
\label{sec:notation}
A design problem is formulated as finding an optimal input or design $\mathbf{x}_\ast \in \mathfrak{X}$ that maximizes the output of an experiment or simulation process $g(\mathbf{x})$,
\begin{eqnarray}
\mathbf{x}_\ast &\triangleq& \underset{\mathbf{x} \in \mathfrak{X}}{\argmax} \ \ g(\mathbf{x}) \ . \label{eq:1}
\end{eqnarray}
As mentioned previously, we cannot access the black-box function $g(\mathbf{x})$ but we are provided with a set $\mathfrak{D}$ of $n$ training data points $(\mathbf{x}_i, z_i)_{i=1}^n$ such that $z_i = g(\mathbf{x}_i)$. This allows us to learn a surrogate $g(\mathbf{x}; \phi)$ of $g(\mathbf{x})$ via supervised learning,
\begin{eqnarray}
\phi &\triangleq& \underset{\phi'}{\argmin} \ \mathcal{L}(\phi'; \mathfrak{D}) \ ,  \label{eq:2}
\end{eqnarray}
where $\mathcal{L}(\phi'; \mathfrak{D}) \ \triangleq\ \sum_{i=1}^n \mathrm{err}\big(g\big(\mathbf{x}_i; \phi'\big),\   z_i\big) $ and
$\phi'$ ($\phi$) denotes the surrogate parameterization and $\mathrm{err}(z', z)$ denotes the loss of predicting $z'$ when the oracle value is $z$. For example, $\mathrm{err}(z', z) = (z' - z)^2$ and $g(\mathbf{x}; \phi) = \phi^\top\mathbf{x}$. Once learned, $\phi$ is fixed and we can use $g(\mathbf{x}; \phi)$ as a surrogate to find (approximately) the optimal design,
\begin{eqnarray}
\mathbf{x}_\phi &\triangleq& \underset{\mathbf{x}\in \mathfrak{X}}{\argmax} \ \ g(\mathbf{x}; \phi) \ . \label{eq:3}  
\end{eqnarray}
The quality of $\mathbf{x}_\phi$ is then defined as the (normalized) difference in oracle output between $\mathbf{x}_\phi$ and the oracle maximizer $\mathbf{x}_\ast$, $\mathfrak{C}(\mathbf{x}_\phi) = | g(\mathbf{x}_\phi) - \min_{\mathbf{x}}g(\mathbf{x})| \ /\ |g(\mathbf{x}_\ast) - \min_{\mathbf{x}}g(\mathbf{x})| \in (0,1)$. Naively, if the surrogate $g(\mathbf{x}; \phi)$ is accurate over the entire input space $\mathfrak{X}$ then solving \eqref{eq:3} is all we need. However, this might only be true near the training data. 

\subsection{Sensitivity of Surrogate Model}
\label{sec:trustability}
Inspired by recent work in assessing model sensitivity \citep{StephensonAISTATS22,TsaiNeurips2021}, the prediction of a model at a particular input $\mathbf{x}$ is considered sensitive if we can find a slightly perturbed variant of it that produces a significantly different prediction at $\mathbf{x}$. Following this intuition, we propose to measure the sensitivity of a model trained on the offline dataset $\mathfrak{D}$ as the probability (over random perturbation) that the absolute difference between its expected output before and after perturbation is larger than a certain threshold. This is formalized below.

\begin{definition} 
\label{def:1}
The $(\alpha, \omega)$-sensitivity of a model $g(\mathbf{x}; \phi)$ on the offline dataset $\mathfrak{D}$ is defined as 
\begin{eqnarray}
\mathcal{S}_\phi\left(\alpha, \omega\right) &\triangleq& \underset{\gamma \sim \mathbb{N}\left(\omega_\mu, \omega_\sigma^2\mathbf{I}\right)}{\mathrm{Pr}}\Big(A(\phi, \gamma) \ \geq\  \alpha\Big),\label{eq:5}
\end{eqnarray}
where $A(\phi,\gamma) \triangleq \left|\mathbb{E}_{\mathbf{x} \sim \mathfrak{D}}\big[g(\mathbf{x}; \phi + \gamma)\big] - \mathbb{E}_{\mathbf{x} \sim \mathfrak{D}}\big[g(\mathbf{x}; \phi)\big]\right|$ and $\omega = (\omega_\mu, \omega_\sigma)$ defines the parameter of the Gaussian distribution of the perturbation $\gamma$.
\end{definition}

Intuitively, the above definition implies that if there is a high chance that the expected output of a model $g(\mathbf{x}; \phi)$ would change significantly (larger than $\alpha$) when its parameterization $\phi$ is perturbed by a certain amount of noise $\gamma$ controlled by $\omega$, then the model is sensitive. Hence, we want to condition the surrogate $g(\mathbf{x}; \phi)$ such that its induced sensitivity $\mathcal{S}_\phi(\alpha, \omega)$ is low. Otherwise, the larger $\mathcal{S}_\phi(\alpha, \omega)$ is, the more likely there exists a neighborhood of input at which the surrogate prediction is brittle against small perturbations of the model. This is established below.

\begin{lemma}
\label{lem:1}
Let $\mathcal{S}_\phi(\alpha, \omega)$ defined via Definition~\ref{def:1}. Suppose $\mathcal{S}_\phi(\alpha, \omega) \geq 1 - \delta$ with $\delta \in (0, 1)$. Then, with probability at least $1 - \delta$ over the space of random perturbation $\gamma \sim \mathbb{N}(\omega_\mu, \omega_\sigma^2\mathbf{I})$, there exists $\mathbf{x} \in \mathfrak{D}$ such that $|g(\mathbf{x}; \phi + \gamma) - g(\mathbf{x}; \phi)| \geq \alpha$ and 
\begin{eqnarray}
\forall \mathbf{x}' \in \mathfrak{D}\ :\ \big\|\mathbf{x}' - \mathbf{x}\big\| \ \leq\  \frac{\alpha}{4\mathfrak{L}_\phi}\ ,\ \text{we have}\ \nonumber \\  \big|g(\mathbf{x}'; \phi + \gamma) - g(\mathbf{x}'; \phi)\big| \ \geq\ \frac{\alpha}{2} \ .
\end{eqnarray}
where $\mathfrak{L}_\phi \triangleq \max_\gamma \mathfrak{L}_{\phi + \gamma}$ and $\mathfrak{L}_{\phi + \gamma}$ denotes the corresponding Lipschitz constant\footnote{$\mathfrak{L}$ is a Lipschitz constant of a function $g(\mathbf{x})$ if it is the smallest value for which $|g(\mathbf{x}) - g(\mathbf{x}')| \leq \mathfrak{L}\|\mathbf{x} - \mathbf{x}'\|$ for all $(\mathbf{x}, \mathbf{x}')$.} of the surrogate $g(.; \phi + \gamma)$. A detailed derivation of this lemma can be found in Appendix~\ref{app:a}.
\end{lemma}

This sensitivity measurement is however relative to a specific configuration of $(\alpha, \omega)$, which needs to be set meticulously. Otherwise, it is easy to see that $\mathcal{S}_\phi(\alpha, \omega)$ would quickly become vacuous (e.g., reaching the maximum value of one regardless of the choice of $\phi$) with decreasing values of $\alpha$ and increasing values of $(\omega_\mu, \omega_\sigma^2)$. Conversely, if $\alpha$ is too large while $(\omega_\mu, \omega_\sigma^2)$ are too small, $\mathcal{S}_\phi(\alpha, \omega)$ instead approaches zero regardless of the choice of $\phi$ and again, becomes vacuous. Intuitively, this means the above sensitivity measure is only meaningful at the right range of values for $\alpha$ and $(\omega_\mu, \omega_\sigma^2)$ which needs to be determined via empirical observations. We suggest a new method to determine the parameters by utilizing only the offline dataset and not accessing the oracle function, described in Appendix~\ref{app:b3}. Furthermore, following the practice in~\citet{TrabuccoICML21}, we use the oracle function to conduct ablation studies in Section~\ref{sec:ablation} to find the most robust, universal values (across all tasks) for $\alpha$ as well as the ranges on which $(\omega_\mu, \omega_\sigma^2)$ are optimized. Interestingly, the parameters  found in Appendix \ref{app:b3} are the same as reported in Section~\ref{sec:ablation}.                                                                 

\subsection{Sensitivity-Informed Regularizer}
\label{sec:regularizer}
The previous discussion suggests that among surrogate candidates $g(\mathbf{x}; \phi)$ that fit equally well to the dataset, we would prefer one whose sensitivity $\mathcal{S}_\phi$ is smallest. Such surrogates tend to have prediction boundaries with (relatively) larger margins, which reduce the risk of being misguided by spurious predictions. Thus, suppose that the surrogate $g(\mathbf{x}; \phi)$ is fitted to offline data via minimizing a loss function $\mathcal{L}(\phi)$, we can regularize it via the following augmentation:
\begin{eqnarray}
    \label{eq:6}
     \phi &=& \argmin_{\phi'} \big(\mathcal{L}(\phi') \ +\ \lambda \cdot \mathcal{S}_{\phi'}(\alpha, \omega)\big) \ ,
\end{eqnarray}
where $\omega \triangleq \argmax_{\omega'} \mathcal{S}_{\phi'}(\alpha, \omega')$ and $\lambda > 0$ is a hyper-parameter regularizing between the two objectives: (1) fitting to offline data and (2) minimizing sensitivity. This features a bi-level optimization task, which can be relaxed into a minimax optimization task,
\begin{eqnarray}
\phi &=& \argmin_{\phi'} \big(\mathcal{L}(\phi') \ +\ \lambda \cdot \max_{\omega'}\mathcal{S}_{\phi'}(\alpha, \omega')\big) \ . \label{eq:7}
\end{eqnarray}
The resulting formulation can now be solved approximately \cuong{via (1) minimizing $\phi'$ and (2) maximizing $\omega'$ simultaneously}. Intuitively, this process features a two-player game where one seeks to decrease the function value while the other seeks to increase it. The optimization thus mimics a fictitious play that often finds an optimal equilibrium between the surrogate and the perturbation model. At this equilibrium, the surrogate has its sensitivity minimized in the worst case (against a most adversarial perturbation). This is in fact similar to the intuition behind Generative Adversarial Networks (GANs) \citep{goodfellow2014generative}.

In addition, our regularization technique here is also agnostic to the specific choice of the loss function $\mathcal{L}(\phi')$. This makes our regularization technique synergistically amenable to a wider range of offline optimizers which could involve both the search and surrogate models. However, the main issue with this approach is that the mathematical characterization of $\mathcal{S}_\phi(\alpha, \omega)$ (see Definition~\ref{def:1}) is not differentiable with respect to either $\omega$ or $\phi$, which prevents it from being optimized with gradient descent/ascent. This is a practical challenge which will be addressed next.\vspace{-3mm}

\section{Practical Algorithm}
\label{sec:algorithm}
To enable numerical optimization of the sensitivity measure $\mathcal{S}_\phi(\alpha, \omega)$ effectively, we will develop approximations of $\mathcal{S}_\phi(\alpha, \omega)$ that can be differentiated with respect to $\phi$ and $\omega$, respectively. These approximations are detailed below.

{\bf Optimizing $\omega$.} First, given the sensitivity threshold $\alpha$ and current surrogate estimate $\phi$, we define
\begin{eqnarray}
\hspace{-6mm}\mathfrak{R}_\alpha(\phi) \hspace{-2mm}&\triangleq&\hspace{-2mm} \left\{\gamma \ \bigg| \ \Big|\mathbb{E}\big[g(\mathbf{x}; \phi + \gamma)\big] -\mathbb{E}\big[g(\mathbf{x}; \phi)\big]\Big| \geq \alpha \right\} \label{eq:8}
\end{eqnarray}
which helps rewrite $\mathcal{S}_\phi(\alpha, \omega) = \mathrm{Pr}(\gamma \in \mathfrak{R}_\alpha(\phi))$ where
\begin{eqnarray}
\hspace{-3mm}\mathrm{Pr}\Big(\gamma \in \mathfrak{R}_\alpha(\phi)\Big) \hspace{-2mm}&=&\hspace{-2mm} \mathbb{E}_\gamma \Big[\mathrm{Pr}\Big(\gamma \in \mathfrak{R}_\alpha(\phi) \Big | \gamma \Big)\Big] \nonumber\\
\hspace{-2mm}&&\hspace{-2mm} \text{with } \gamma \sim \mathbb{N}(\omega_\mu, \omega^2_\sigma \mathbf{I})\nonumber\\
\hspace{-2mm}&=&\hspace{-2mm} \mathbb{E}_\epsilon \Big[\mathrm{Pr}\Big(\omega_\mu + \omega_\sigma \cdot \epsilon  \in \mathfrak{R}_\alpha(\phi) \Big | \epsilon \Big)\Big], \nonumber\\ 
\hspace{-2mm}&&\hspace{-2mm} \text{with } \epsilon \sim \mathbb{N}(0,\mathbf{I}) \nonumber\\ 
\hspace{-2mm}&\simeq&\hspace{-2mm} \frac{1}{m}\sum_{i=1}^m\Phi\Big(\gamma_i; \mathbf{w}\Big)\label{eq:9}
\end{eqnarray}
where $\gamma_i \triangleq \omega_\mu + \omega_\sigma^2 \cdot \epsilon_i$ and $\{\epsilon_i\}_{i=1}^m$ are identical and independent samples drawn from $\mathbb{N}(0, \mathbf{I})$ while $\Phi(\gamma_i; \mathbf{w})$ is a learnable neural net that predicts the probability that $\gamma_i \triangleq \omega_\mu + \omega_\sigma^2 \cdot \epsilon_i \in \mathfrak{R}_\alpha(\phi)$. 

We will refer to the last step in \eqref{eq:9} above as the neural re-parameterization, which can approximate $\mathcal{S}_\phi(\alpha, \omega)$ arbitrarily closely given a sufficiently large number $m$ of samples and that the parameterization of $\Phi(\gamma; \mathbf{w})$ is sufficiently rich. Given a set of samples $(\gamma_i, \kappa_i)_{i=1}^m$ where $\kappa_i \triangleq \mathbb{I}(\gamma_i \in \mathfrak{R}_\alpha(\phi)) \in \{0, 1\}$, we can learn this neural re-parameterization via solving
\begin{eqnarray}
\hspace{-15mm}\mathbf{w} \hspace{-2mm}&=&\hspace{-2mm} \argmax_{\mathbf{w}'}\ \   \bigg[\Big(1 - \kappa_i\Big)\log\Big( 1- \Phi\Big(\gamma_i; \mathbf{w}'\Big)\Big) \nonumber \\ 
\hspace{-2mm}&&\hspace{-2mm} \ \ \ \ \ \ \ \ \ \ \ \ \ \ \ \ \ + \ \kappa_i\log\Phi\Big(\gamma_i; \mathbf{w}'\Big)\bigg]    \label{eq:10}
\end{eqnarray}
which is a standard logistic regression losses with parameterized bias $\Phi(\gamma;\mathbf{w})$ and training data $(\gamma_i, \kappa_i)^m_{i=1}$. Note that $\kappa_i = \mathbb{I}(\gamma_i \in \mathfrak{R}_\alpha(\phi))$ which is tractable. Once learned, we can take advantage of the differentiability of $\Phi(\gamma; \mathbf{w})$ to compute the gradient of $\mathcal{S}_\phi$ with respect to $\omega$:
\begin{eqnarray}
\hspace{-12mm}\frac{\partial\mathcal{S}_\phi}{\partial\omega} &\simeq& \frac{1}{m}\sum_{i=1}^m \frac{\partial \Phi}{\partial\omega} \ \ =\ \ \frac{1}{m}\sum_{i=1}^m \left(\frac{\partial\Phi}{\partial\gamma_i} \cdot \frac{\partial\gamma_i}{\partial\omega}\right) \label{eq:11}
\end{eqnarray}
Now recall that by the change of parameter, $\gamma_i = \omega_\mu + \omega_\sigma \cdot \epsilon_i$. Thus, $\partial\gamma_i / \partial\omega_\mu = 1$ and $\partial\gamma_i /\partial\omega_\sigma = \epsilon_i$. As such, replacing $\omega$, respectively, with $\omega_\mu$ and $\omega_\sigma$ in \eqref{eq:11} produces\footnote{Note that we abuse the notation a bit here to treat $\epsilon_i$ as a scalar while it should be a random vector. This is not an issue since the covariance matrix of its distribution $\mathbb{N}(0, \mathbf{I})$ is diagonal, which implies components of the random vector are i.i.d and hence, \eqref{eq:12} applies separately to each such scalar component.}
\begin{eqnarray}
\hspace{-3mm}\frac{\partial\mathcal{S}_\phi}{\partial\omega_\mu} = \frac{1}{m} \sum_{i=1}^m \frac{\partial\Phi}{\partial\gamma_i} \quad \text{and} \quad \frac{\partial\mathcal{S}_\phi}{\partial\omega_\sigma} = \frac{1}{m} \sum_{i=1}^m \epsilon_i \cdot \frac{\partial\Phi}{\partial\gamma_i} \label{eq:12}
\end{eqnarray}
which enables maximization of $\omega$ via gradient ascent.

\begin{algorithm}[t]
\small
\caption{\ourmethod~}\label{alg:pseudo-code}
\textbf{Input:} offline data \(\mathfrak{D} = \{(\mathbf{x}_i,z_i)\}_{i=1}^n\); initial surrogate model \(g(\mathbf{x};\phi)\); initial perturbation parameters \(\omega = (\omega_\mu, \omega_\sigma^2)\); no. $m$ of sampled perturbations $\gamma \sim \mathbb{N}(\omega_\mu, \omega_\sigma^2\mathbf{I})$; no. of iterations \(\tau\); sensitivity threshold \(\alpha\); learning rates \((\eta_\omega, \eta_\phi)\); Bound of $\omega_\mu$ [$\omega_{\mu_l}, \omega_{\mu_u}$]; Bound of $\omega_\sigma$ [$\omega_{\sigma_l}, \omega_{\sigma_u}$]. \\
\begin{algorithmic}[1]
\STATE Initialize $\phi^{(1)} \leftarrow \phi$ and $\omega^{(1)} \leftarrow \omega$
\FOR{\(t \gets 1 : \tau\)}
    \STATE Sample $\left\{\gamma_i\right\}_{i=1}^m$ : $\gamma_i =\omega^{(t)}_\mu + \omega^{(t)}_\sigma \cdot \epsilon_i$ with $\epsilon_i \sim \mathbb{N}(0, \mathbf{I})$ \\
    \STATE $\nabla_\phi h(\phi) = \mathbb{E}_{\mathbf{x} \sim \mathfrak{D}}\Big[\nabla_\phi g(\mathbf{x};\phi)\Big]$ at $\phi = \phi^{(t)}$ -- see Eq.~\eqref{eq:13}
    
    \FOR{\(i \gets 1 : m\)}
        \STATE \(\kappa_i \gets \mathbb{I}\left({\Big|\nabla_\phi h(\phi)^\top \gamma_i\Big|>\alpha}\right)\) -- see Eq.~\ref{eq:8}
    \ENDFOR 
    
    \STATE Learn \(\Phi(\gamma;\mathbf{w})\) with dataset \(\{(\gamma_i,\kappa_i)\}_{i=1}^m\) -- see Eq.~\ref{eq:10}

    \STATE {\bf \underline{Optimizing $\omega$:}}
    
    \STATE  $\nabla_{\omega_\mu}\mathcal{S}_\phi \leftarrow m^{-1} \Big(\sum_{i=1}^m \nabla_\gamma\Phi(\gamma_i)\Big)$ via Eq.~\ref{eq:12}
    
    \STATE $\nabla_{\omega_\sigma}\mathcal{S}_\phi \leftarrow m^{-1} \Big(\sum_{i=1}^m\epsilon_i \cdot  \nabla_\gamma\Phi(\gamma_i)\Big)$ via Eq.~\ref{eq:12}
    \STATE   
    \(\omega^{(t + 1)}_\mu \gets \omega^{(t)}_\mu + \eta_\omega \cdot \nabla_{\omega_\mu}\mathcal{S}_\phi\) \STATE 
    \(\omega^{(t + 1)}_\sigma \gets \omega^{(t)}_\sigma + \eta_\omega \cdot \nabla_{\omega_\sigma}\mathcal{S}_\phi\)
    \STATE \cuong{Project $\omega^{(t + 1)}_\mu$ into [$\omega_{\mu_l}, \omega_{\mu_u}$]}
    \STATE \cuong{Project $\omega^{(t + 1)}_\sigma$ into [$\omega_{\sigma_l}, \omega_{\sigma_u}$]}
    \STATE {\bf \underline{Optimizing $\phi$:}}
    
    \STATE \(\mathcal{S}^{+}_{\phi}(\alpha,\omega) = m^{-1}\sum_{i=1}^m \left[ \min \left(1, \left(\left(\nabla_\phi h(\phi)^\top \gamma_i\right)^2 / \alpha^2\right) \right) \right]\) see Eq.~\ref{eq:12a} and Eq.~\ref{eq:13}

    \STATE \(\phi^{(t+1)} \leftarrow \phi^{(t)} + \eta_\phi \cdot \Big(\nabla_\phi\mathcal{L}(\phi; \mathfrak{D}) + \lambda \cdot \nabla_{\phi}\mathcal{S}^+_\phi(\alpha, \omega)\Big)\) at $\phi = \phi^{(t)}$ and $\omega = \omega^{(t + 1)}$    
\ENDFOR
\STATE \textbf{return} learned surrogate \(g\left(\mathbf{x}; \phi^{(\tau + 1)}\right)\)\\
\end{algorithmic}
\end{algorithm}

{\bf Optimizing $\phi$.} To optimize $\phi$, we need another approximation since the neural re-parameterization above is still not differentiable with respect to $\phi$. This is developed via the following lemma.

\begin{lemma}
\label{lem:2}
Let $\mathcal{S}_\phi(\alpha, \omega)$ defined via Definition~\ref{def:1}. We have $\mathcal{S}_\phi(\alpha, \omega) \leq \mathcal{S}^+_\phi(\alpha, \omega)$ where
\begin{eqnarray}
\hspace{-17mm}\mathcal{S}^+_\phi(\alpha, \omega) &=& \mathbb{E}_\gamma \left[\min\Big(1, (A(\phi, \gamma)/\alpha)^2\Big)\right]
\label{eq:12a}
\end{eqnarray}
where $\gamma \sim \mathbb{N}(\omega_\mu, \omega_\sigma^2\mathbf{I})$ and $A(\phi,\gamma)$ is defined previously in Definition~\ref{def:1}. A detailed derivation of this lemma is provided in Appendix~\ref{app:c}.
\end{lemma}

Lemma~\ref{lem:2} establishes an upper bound for $\mathcal{S}_\phi(\alpha, \omega)$, which is now differentiable with respect to $\phi$. Note that while minimizing $\phi$, $\omega = (\omega_\mu, \omega_\sigma^2)$ is fixed and hence, the expectation does not involve parameters that need to be differentiated. Thus, the derivative operator can be pushed inside the expectation, which is differentiable with respect to $\phi$. 

However, one minor detail here is that computing $\mathcal{S}^+_\phi(\alpha, \omega)$ still requires computing the expected output difference for each sampled $\gamma$. This operation is not vectorizable and cannot take advantage of the GPU compute infrastructure. To sidestep this final hurdle, we propose to approximate $h(\phi + \gamma) \triangleq \mathbb{E}[g(\mathbf{x}; \phi + \gamma)]$ with a first-order Taylor expansion around $\phi$. That is,
$\mathbb{E}[g(\mathbf{x}; \phi + \gamma)] - \mathbb{E}[g(\mathbf{x}; \phi)] = $
\begin{eqnarray}
&& h(\phi + \gamma) - h(\phi) \ \ \simeq\ \ \nabla h(\phi)^\top \gamma \label{eq:13}
\end{eqnarray}
which is computable via a (vectorizable) matrix multiplication involving $\gamma$. For clarity, a full pseudo-code of our method is detailed in Algorithm~\ref{alg:pseudo-code}.

{\bf Remark.} As an alternative approach, the Taylor approximation presented in Eq.~\eqref{eq:13} can also be applied directly to $(A(\phi, \gamma)$ in Eq.~\eqref{eq:5}. This results in an approximation of $\mathcal{S}_\phi(\alpha, \omega)$ in terms of the Gaussian cumulative distribution function (CDF) detailed in Appendix~\ref{sec:SIRO-CDF}, which is differentiable with respect to all parameters. Consequently, Eq.~\eqref{eq:5} can be optimized directly without using the relaxation form in Lemma~\ref{lem:2} or the use of the neural re-parameterization in Eq.~\eqref{eq:9}. This alternative approximation of the proposed regularizer also helps improve well over the baseline performance but the overall improvement is less pronounced than that of \ourmethod, as reported in Appendix~\ref{sec:SIRO-CDF}.

\section{Experiments}
\label{sec:experiment}
This section evaluates the effectiveness of our proposed regularizer (\ourmethod) on boosting the performance of existing state-of-the-art offline optimizers. Our empirical studies adopt the benchmark tasks from Design-Bench \citep{trabucco2022design} and its widely recognized baseline algorithms and evaluation protocol (Section~\ref{sec:benchmark}). All empirical results and analyses are reported in Section~\ref{sec:results}.

\subsection{Benchmarks, Baselines and Evaluation}
\label{sec:benchmark}

{\bf Benchmark Tasks.} Our empirical evaluations are conducted on $6$ real-world tasks from Design-Bench \citep{trabucco2022design}, with both discrete and continuous domains. 

The discrete tasks include \textbf{TF-Bind-8}, \textbf{TF-Bind-10}, and \textbf{ChEMBL} where {\bf TF-Bind-8} and {\bf TF-Bind-10} \citep{tfbind8} involve discovering DNA sequences with high binding affinity to a specific transcription factor (SIX6 REF R1) with sequence lengths $8$ and $10$, respectively; and \textbf{ChEMBL} is derived from a drug property database \citep{gaulton2012chembl} and requires optimizing a molecule for a high MCHC value when paired with assay CHEMBL3885882.

The continuous tasks include \textbf{Ant Morphology} \citep{brockman2016openai}, \textbf{D'Kitty Morphology} \citep{ahn2020robel}, and \textbf{Superconductors} \citep{BrookeICML19}. In \textbf{Ant Morphology} and the \textbf{D'Kitty Morphology} task, we optimize the physical structure of a simulated robot Ant from OpenAI Gym \citep{brockman2016openai} and the D’Kitty robot from ROBEL \citep{ahn2020robel}. The \textbf{Superconductor} task is about designing superconductor molecules that have the highest critical temperature. 

{\bf Baselines.} To assess how our proposed regularizer \ourmethod~influences the performance of established baseline algorithms, we selected $11$ widely recognized offline optimizers for comparison. These include \textbf{BO-qEI} \citep{trabucco2022design}, \textbf{CbAS} \citep{BrookeICML19}, 
\textbf{RoMA} \citep{YuNeurIPs21}, \textbf{ICT} \citep{yuan2023importance}, \textbf{CMA-ES} \citep{hansen2006cma}, \textbf{COMs} \citep{TrabuccoICML21}, \textbf{MINs} \citep{KumarNeurIPs20}, \textbf{REINFORCE} \citep{williams1992simple}, and $3$ variants of gradient ascent (\textbf{GA}, \textbf{ENS-MIN}, \textbf{ENS-MEAN}) which correspond to the vanilla gradient ascent, the min ensemble of gradient ascent, and the mean ensemble of gradient ascent.

{\bf Evaluation Protocol.} For each baseline algorithm, we configure it with its corresponding best hyperparameters specified in \citep{trabucco2022design}. To provide a comprehensive evaluation of each algorithm's performance, we adhere to the recommended approach in \citep{trabucco2022design}, which requires each method to generate \(K = 128\) optimized design candidates, which are then evaluated using the oracle function. The evaluated performance of these candidates are then sorted in increasing order from which performance at $50$-th and $75$-th and $100$-th percentile levels are reported. All reported performance are averaged over $8$ independent runs. 


{\bf Hyper-parameter Configuration.} Our proposed regularizer \ourmethod~also has additional hyperparameters $(\alpha, \omega)$ as highlighted previously in Section~\ref{sec:trustability}. In particular, $\omega = (\omega_\mu, \omega_\sigma^2)$ is a tuple of learnable parameters that define the (adversarial) perturbation distribution \(\mathbb{N}(\omega_\mu\mathbf{1}, \omega_\sigma^2\mathbf{I})\), which is used to measure the sensitivity of the regularizer. However, the perturbation is supposed to be in the low-noise regime which requires the range of values for those parameters to be set so that the perturbation will not dominate the surrogate's parameters. Otherwise, our sensitivity measure will become vacuous. We have conducted ablation studies in Section~\ref{sec:ablation} to find the most appropriate bounds for these parameters, which appear to be \cuong{$[-10^{-3}, 10^{-3}]$ for $[\omega_{\mu_l}, \omega_{\mu_u}]$ and $[10^{-5}, 10^{-2}]$ for $[\omega_{\sigma_l}, \omega_{\sigma_u}]$}. We use the above bounds in all experiments, in which $\omega_\mu$ and $\omega_\sigma^2$ are initialized to $0$ and $10^{-3}$, respectively. Likewise, for the sensitivity threshold, our ablation studies observe the impact of several values of $\alpha$ on the performance and find that $\alpha = 0.1$ is the best universal value for \ourmethod. In addition, we set the weight of the regularizer \(\lambda\) to \(10^{-3}\), the no. $m$ of perturbation sample per iteration to \(100\) and the learning rates $\eta_{\omega} = 10^{-2}$, $\eta_{\phi} = 10^{-3}$. We find that this configuration is universally robust across all benchmark tasks. $\Phi$ is a neural network with one hidden layer comprising two hidden units and one output layer.


\subsection{Results and Discussion}
\label{sec:results}
This section reported the percentage of improvement over baseline performance achieved by \ourmethod~when it is applied to an existing baseline. We have evaluated this at $50$-th, $75$-th and $100$-th percentile levels. However, due to limited space, we only report result of the $100$-th percentile level in the main text. The other results are instead deferred to Appendix~\ref{app:75_50_percentile}.

\begin{table*}[h]
    \caption{Percentage of performance improvement achieved by \ourmethod~across all tasks and baselines at the \textbf{$100$-th percentile} level. \textbf{P} denote the achieved normalized performance while \textbf{G} denote \ourmethod's percentage of gain over baseline performance.}
    \centering
    \setlength\tabcolsep{3pt} 
    \resizebox{\textwidth}{!}{
        \begin{tabular}{ll|ll|ll|ll|ll|ll|ll} 
         \toprule
           & & \multicolumn{6}{c|}{\textbf{Continuous Tasks}} & \multicolumn{6}{c}{\textbf{Discrete Tasks}} 
           \\\cmidrule(lr){3-8}\cmidrule(lr){9-14}
           & & \multicolumn{2}{c}{\textbf{Ant Morphology}} & \multicolumn{2}{c}{\textbf{D'Kitty Morphology}} & \multicolumn{2}{c|}{\textbf{Superconductor}} 
          & \multicolumn{2}{c}{\textbf{TF Bind 8}} & \multicolumn{2}{c}{\textbf{TF Bind 10}} & \multicolumn{2}{c}{\textbf{ChEMBL}}
          
          \\ \cmidrule(lr){3-4}\cmidrule(lr){5-6} \cmidrule(lr){7-8}\cmidrule(lr){9-10}\cmidrule(lr){11-12}\cmidrule(lr){13-14}

          \multicolumn{2}{c|}{\textbf{Algorithms}}& \multicolumn{1}{c}{\textbf{P}} & \multicolumn{1}{c}{\textbf{G}}
           & \multicolumn{1}{c}{\textbf{P}} & \multicolumn{1}{c}{\textbf{G}}
           & \multicolumn{1}{c}{\textbf{P}} & \multicolumn{1}{c|}{\textbf{G}}
           & \multicolumn{1}{c}{\textbf{P}} & \multicolumn{1}{c}{\textbf{G}}
           & \multicolumn{1}{c}{\textbf{P}} & \multicolumn{1}{c}{\textbf{G}}
           & \multicolumn{1}{c}{\textbf{P}} & \multicolumn{1}{c}{\textbf{G}}
          \\ \midrule 
         \(\mathfrak{D}\)(\textbf{best}) & & 0.565 & & 0.884 & & 0.400 & & 0.439 & & 0.467 & & 0.605
         \\ \midrule
         
         \multirow{2}{*}{\textbf{CbAS}} & Base 
         & 0.856 ± 0.029 & 
         & 0.895 ± 0.011 & 
         & 0.480 ± 0.038 & 
         & 0.911 ± 0.034 & 
         & 0.615 ± 0.031 & 
         & 0.636 ± 0.005 & \\
         & \ourmethod 
         & 0.861 ± 0.032 & \color{blue}+0.5\% 
         & 0.907 ± 0.012 & \color{blue}+1.2\% 
         & 0.485 ± 0.025 & \color{blue}+0.5\% 
         & 0.919 ± 0.054 & \color{blue}+0.8\% 
         & 0.656 ± 0.042 & \color{blue}+4.1\%
         & 0.637 ± 0.010 & \color{blue}+0.1\%
         \\ \midrule
         
        \multirow{2}{*}{ \textbf{BO-qEI}} & Base 
        & 0.812 ± 0.000 &
        & 0.896 ± 0.000 &
        & 0.394 ± 0.048 &
        & 0.779 ± 0.125 & 
        & 0.692 ± 0.126 &
        & 0.659 ± 0.023 &\\
         & \ourmethod
         & 0.812 ± 0.000 & \color{blue}+0.0\% 
         & 0.896 ± 0.000 & \color{blue}+0.0\%
         & 0.464 ± 0.013 & \color{blue}+7.0\% 
         & 0.802 ± 0.069 & \color{blue}+2.3\% 
         & 0.692 ± 0.126 & \color{blue}+0.0\%
         & 0.688 ± 0.000 & \color{blue}+2.9\%
         \\ \midrule
         
         \multirow{2}{*}{\textbf{CMA-ES}} & Base 
         & 1.915 ± 0.909 &
         & 0.723 ± 0.001 &
         & 0.481 ± 0.026 &
         & 0.944 ± 0.035  &
         & 0.676 ± 0.039 &
         & 0.633 ± 0.000 &\\
         & \ourmethod 
         & 2.009 ± 1.540 & \color{blue}+9.4\% 
         & 0.725 ± 0.002 & \color{blue}+0.2\% 
         & 0.482 ± 0.024 & \color{blue}+0.1\% 
         & 0.941 ± 0.029 & \color{red}-0.3\% 
         & 0.672 ± 0.063 & \color{red}-0.4\%
         & 0.633 ± 0.000 & \color{blue}+0.0\%
         \\ \midrule
         
         \multirow{2}{*}{\textbf{GA}} & Base 
         & 0.299 ± 0.037 & 
         & 0.871 ± 0.012 & 
         & 0.506 ± 0.008 & 
         & 0.980 ± 0.015 & 
         & 0.647 ± 0.029 & 
         & 0.640 ± 0.010 & \\
         & \ourmethod 
         & 0.314 ± 0.034 & \color{blue}+1.5\% 
         & 0.883 ± 0.012 & \color{blue}+1.2\% 
         & 0.515 ± 0.014 & \color{blue}+0.9\% 
         & 0.986 ± 0.007 & \color{blue}+0.6\% 
         & 0.658 ± 0.072 & \color{blue}+1.1\% 
         & 0.646 ± 0.002 & \color{blue}+0.6\%
         \\ \midrule
         
         \multirow{2}{*}{\textbf{ENS-MIN}} & Base 
         & 0.399 ± 0.077 & 
         & 0.892 ± 0.010 & 
         & 0.501 ± 0.013 & 
         & 0.986 ± 0.006 & 
         & 0.642 ± 0.025 & 
         & 0.653 ± 0.018 & \\
         & \ourmethod  
         & 0.472 ± 0.110 &  \color{blue}+7.3\% 
         & 0.893 ± 0.009 & \color{blue}+0.1\% 
         & 0.504 ± 0.010 & \color{blue}+0.3\% 
         & 0.989 ± 0.007 & \color{blue}+0.3\% 
         & 0.662 ± 0.038 & \color{blue}+2.0\%
         & 0.662 ± 0.007 & \color{blue}+0.9\%
         \\ \midrule
         
         \multirow{2}{*}{\textbf{ENS-MEAN}} & Base 
         & 0.403 ± 0.045 & 
         & 0.897 ± 0.009 & 
         & 0.510 ± 0.012 & 
         & 0.984 ± 0.007 & 
         & 0.628 ± 0.028 & 
         & 0.653 ± 0.014 & \\
         & \ourmethod  
         & 0.412 ± 0.094 & \color{blue}+0.9\% 
         & 0.897 ± 0.005 & \color{blue}+0.0\% 
         & 0.511 ± 0.015 & \color{blue}+0.1\% 
         & 0.986 ± 0.007 & \color{blue}+0.2\% 
         & 0.641 ± 0.032 & \color{blue}+1.3\%
         & 0.662 ± 0.009 & \color{blue}+0.9\%
         \\ \midrule
         
         \multirow{2}{*}{\textbf{REINFORCE}} & Base 
         & 0.253 ± 0.047 & 
         & 0.674 ± 0.138 & 
         & 0.481 ± 0.015 & 
         & 0.929 ± 0.031 & 
         & 0.664 ± 0.061 & 
         & 0.634 ± 0.002 & \\
         & \ourmethod 
         & 0.278 ± 0.014 & \color{blue}+2.5\% 
         & 0.732 ± 0.003 & \color{blue}+5.8\% 
         & 0.483 ± 0.007 & \color{blue}+0.2\% 
         & 0.943 ± 0.029 & \color{blue}+1.4\% 
         & 0.964 ± 0.096 & \color{blue}+30.0\%
         & 0.639 ± 0.010 & \color{blue}+0.5\%
         \\ \midrule
         
        \multirow{2}{*}{ \textbf{MINs}} & Base 
        & 0.906 ± 0.019 & 
        & 0.944 ± 0.009 & 
        & 0.461 ± 0.027 & 
        & 0.907 ± 0.051 & 
        & 0.636 ± 0.039 & 
        & 0.633 ± 0.000 & \\
         & \ourmethod  
         & 0.924 ± 0.016 & \color{blue}+1.8\% 
         & 0.942 ± 0.008 & \color{red}-0.2\% 
         & 0.476 ± 0.023 & \color{blue}+1.5\% 
         & 0.938 ± 0.053 & \color{blue}+3.1\% 
         & 0.644 ± 0.047 & \color{blue}+0.8\%
         & 0.634 ± 0.003 & \color{blue}+0.1\%
         \\ \midrule
         
         \multirow{2}{*}{\textbf{COMs}} & Base 
         & 0.896 ± 0.024 & 
         & 0.937 ± 0.012 & 
         & 0.483 ± 0.026 & 
         & 0.946 ± 0.035 & 
         & 0.628 ± 0.044 & 
         & 0.633 ± 0.000 & \\
        & \ourmethod  
        & 0.918 ± 0.025 & \color{blue}+2.2\% 
        & 0.942 ± 0.011 & \color{blue}+0.5\% 
        & 0.486 ± 0.030 & \color{blue}+0.3\% 
        & 0.954 ± 0.020 & \color{blue}+0.8\% 
        & 0.664 ± 0.038 & \color{blue}+3.6\%
        & 0.638 ± 0.009 & \color{blue}+0.5\%
        \\ \midrule

        \multirow{2}{*}{\textbf{RoMA}} & Base 
         & 0.574 ± 0.073 &
         & 0.821 ± 0.019 &
         & 0.490 ± 0.022 & 
         & 0.665 ± 0.000 & 
         & 0.547 ± 0.011 & 
         & 0.633 ± 0.000 & \\
        & \ourmethod 
        & 0.607 ± 0.087 & \color{blue}+3.4\% 
        & 0.830 ± 0.028 & \color{blue}+0.9\% 
        & 0.504 ± 0.022 & \color{blue}+1.4\% 
        & 0.665 ± 0.000 & \color{blue}+0.0\% 
        & 0.553 ± 0.000 & \color{blue}+0.6\%
        & 0.633 ± 0.000 & \color{blue}+0.0\%
        \\ \midrule

        \multirow{2}{*}{\textbf{ICT}} & Base 
         & 0.930 ± 0.030 &
         & 0.938 ± 0.012 &
         & 0.489 ± 0.018 &
         & 0.911 ± 0.049 &
         & 0.655 ± 0.022 &
         & 0.633 ± 0.000 & \\
        & \ourmethod  
        & 0.939 ± 0.013 & \color{blue}+0.9\% 
        & 0.943 ± 0.013 & \color{blue}+0.5\% 
        & 0.501 ± 0.022 & \color{blue}+1.2\% 
        & 0.918 ± 0.023 & \color{blue}+0.7\% 
        & 0.667 ± 0.033 & \color{blue}+1.2\%
        & 0.643 ± 0.017 & \color{blue}+1.0\%
        \\\bottomrule
        \end{tabular}
    }
    \label{tb:100th_results}
\end{table*}

{\bf Results on Continuous Tasks.}
The first part (the first $3$ column) of Table~\ref{tb:100th_results} shows that among $33$ cases (across $11$ baselines and $3$ tasks), incorporating the \ourmethod~regularizer improves the baseline performance positively up to $9.4\%$. There is only one instance where \ourmethod~decreases the performance but the decrease is only $0.2\%$, which is negligible. 

Moreover, in some cases, even when \ourmethod~only maintains similar performance as the baseline, it still helps reduce the performance variance, as demonstrated by a reduction from 0.9\% to 0.5\% with {\bf ENS-MEAN} baseline on the \textbf{D'kitty Morphology} task. In addition, for certain cases, \ourmethod~also helps establish new SOTA performance. For example, in the \textbf{Ant Morphology} task where the SOTA baseline is represented by \textbf{CMA-ES} achieving 191.5\%, incorporating \ourmethod~elevates its performance by $200.9\%$, setting a new SOTA performance. 

{\bf Results on Discrete Tasks.}
The second part (last $3$ columns) of Table~\ref{tb:100th_results} illustrates the impact of our regularizer \ourmethod~ on the performance of baseline algorithms in $3$ discrete domains (\textbf{TF-BIND-8}, \textbf{TF-BIND-10}, and \textbf{ChEMBL}). It is observed that in most cases ($28/33$), \ourmethod~enhances the baseline algorithms' performance significantly up to $30\%$. 

For the two instances in which \ourmethod~decreases the performance, the amount of decrease is (relatively) much milder, ranging between $0.2\%$ and $0.4\%$. Furthermore, on a closer look, it is also observable that for certain cases, the incorporation of \ourmethod~also helps elevate the state-of-the-art (SOTA) results (in addition to improving the baseline performance). For example, on {\bf TF-BIND-8} and {\bf ChEMBL}, the original SOTA performance ($0.986$ and $0.659$) are achieved by \textbf{ENS-MIN} and \textbf{BO-qEI}, respectively. These are further elevated to $0.989$ and $0.688$ when \ourmethod~is added to regulate the loss function of \textbf{ENS-MIN} and \textbf{BO-qEI}, establishing new SOTA. Overall, our observations suggest that \ourmethod~ demonstrates consistently a high probability ($84.85\%$ = $56/66$ cases) of improving baseline performance. On average, it leads to an improvement of approximately $2.08\%$, with a notable peak improvement of $30\%$. Conversely, \ourmethod~also carries a relatively much lower probability ($4.5\%$ = $3/66$ cases) of decreasing baseline performance. When such cases occur, the average performance decrease is at most $0.3\%$, which is almost negligible. The code for reproducing our results is at
\url{https://github.com/daomanhcuonghust/BOSS}

\subsection{Ablation Experiments}
\label{sec:ablation}
\begin{figure*}
     \centering
     \small
     \hfill
     \begin{minipage}{0.48\linewidth} 
        \centering
        \includegraphics[width=\textwidth]{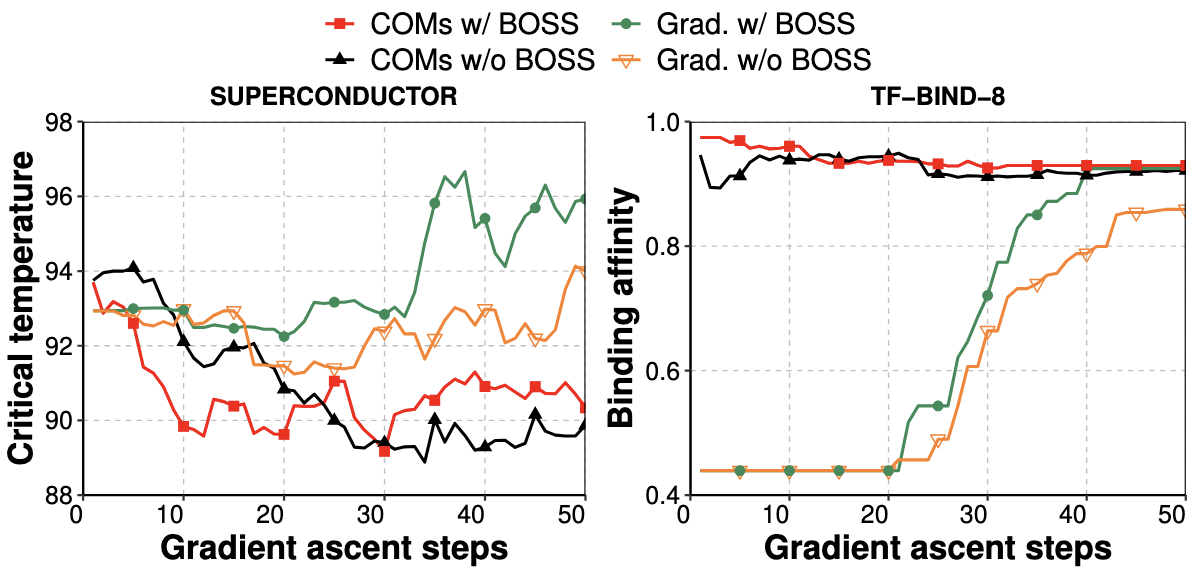}\\
        (a)
         \label{fig:compare_coms_grad}
     \end{minipage}
     \hfill
     \begin{minipage}{0.48\linewidth} 
        \centering
        \includegraphics[width=\textwidth]{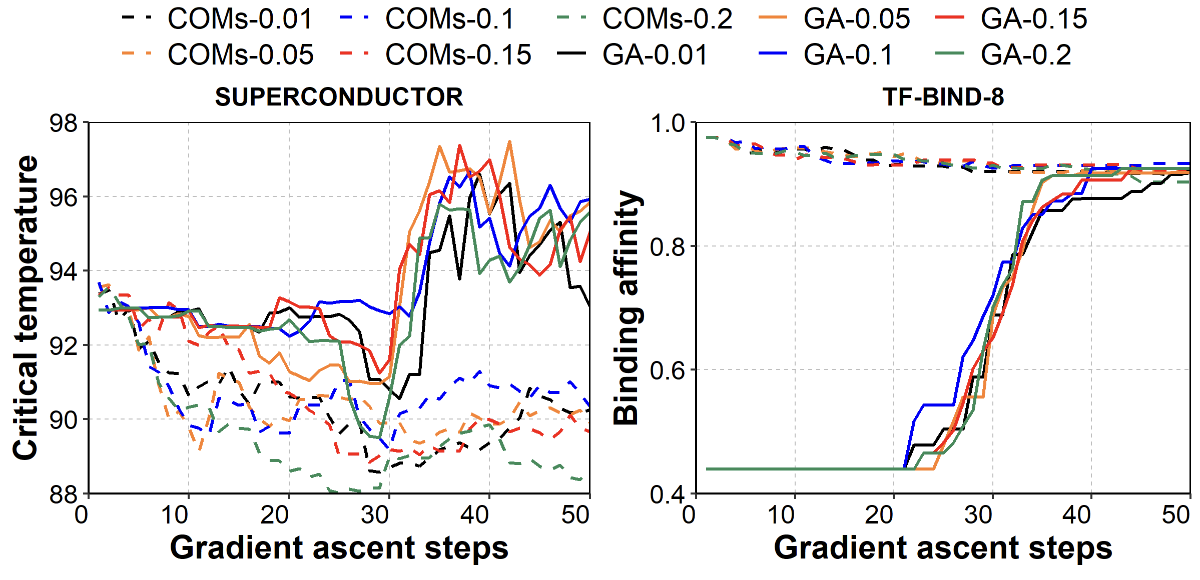}\\
        (b)
         \label{fig:change_alpha}
     \end{minipage}
     \hfill\\
     \hfill
     \begin{minipage}{0.48\linewidth} 
        \centering
        \includegraphics[width=\textwidth]{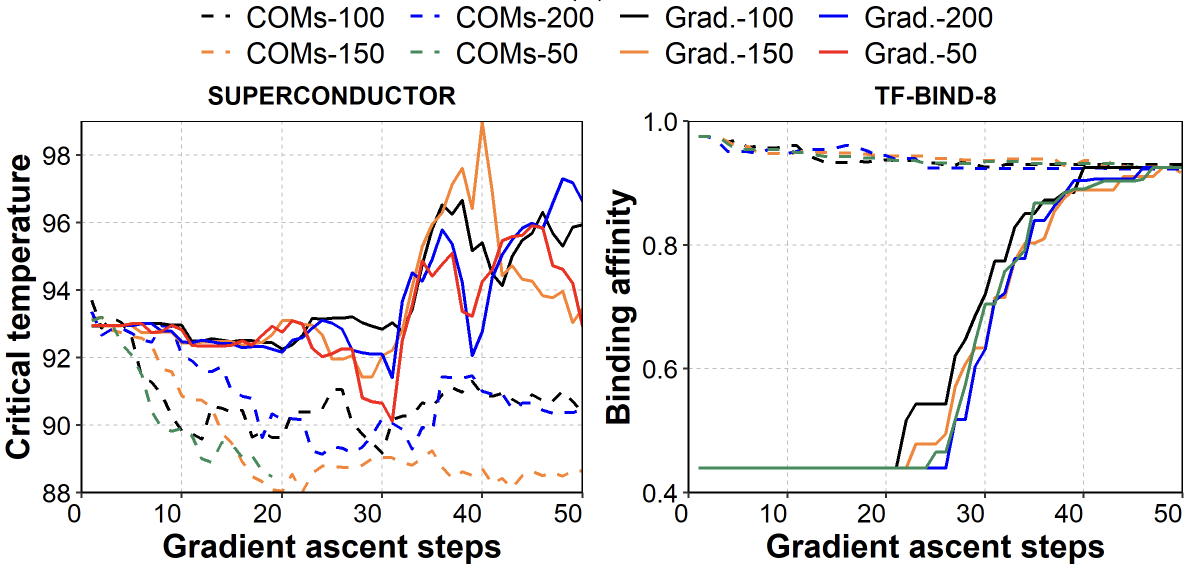}\\
        (c)
         \label{fig:change_m}
     \end{minipage}
     \hfill
     \begin{minipage}{0.48\linewidth} 
        \centering
        \includegraphics[width=\textwidth]{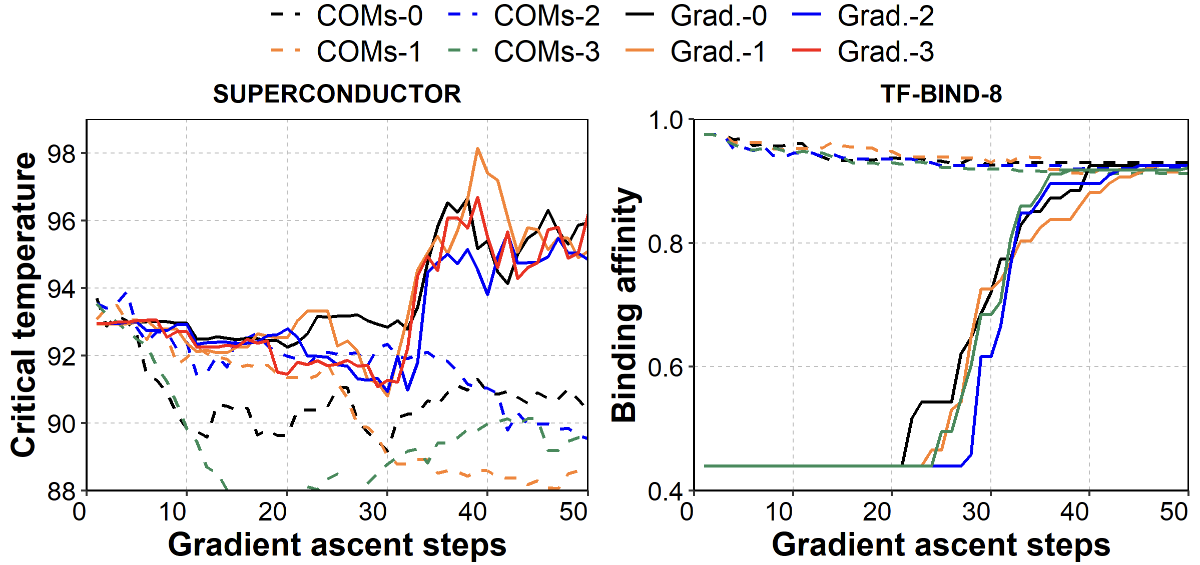}\\
        (d)
         \label{fig:change_omega_bound}
     \end{minipage}
     \caption{Plots of performance variation of \textbf{COMS} and \textbf{GA} (regularized by \ourmethod) to changes in (a) the no. of gradient ascent steps during optimization; (b) values of the sensitivity threshold $\alpha$; (c) changes in the no. of perturbation samples $m$; and (d) changes in value ranges of bound \(([\omega_{\mu_l},\omega_{\mu_u}],[\omega_{\sigma_l},\omega_{\sigma_u}])\) for parameter $\omega$ of the perturbation distribution. These are \(([-10^{-3},10^{-3}],[10^{-5},10^{-2}])\), \( ([-10^{-3},10^{-3}],[10^{-6},10^{-3}])\), \( ([-10^{-2},10^{-2}],[10^{-5},10^{-2}])\), and \( ([-10^{-2},10^{-2}],[10^{-6},10^{-3}])\) which are indexed with 0, 1, 2, 3 in this figure.}
     \label{fig:ablation}
\end{figure*}
This section presents additional experiments to examine the sensitivity of two representative baselines, {\bf COMs} and {\bf GA} (regularized with \ourmethod) to changes in the number of gradient ascent steps performed during optimization. Furthermore, we also conduct ablation studies to investigate the effects of specific hyperparameters of \ourmethod~(see Algorithm~\ref{alg:pseudo-code}), including the sensitivity threshold \(\alpha\), the no. of sampled perturbations \(m\), and the effective value ranges for \(\omega_\mu, \omega_\sigma^2\), on the performance. Our studies are mainly conducted on two tasks: \textbf{SUPERCONDUCTOR} and \textbf{TF-BIND-8}. \cuong{In addition, we also present empirical experiments to demonstrate the tightness of Lemma \ref{lem:2}, validating the localized conditioning effect with relevant metrics and computational complexity. Finally, we present tuning of $\lambda$, convergence analysis, and limitations in Appendices \ref{app:tuning_lambda}, \ref{app:convergence_analysis}, and \ref{app:limitation}, respectively, due to limited space.}

{\bf \ourmethod~enhances stability of {\bf COMs} and gradient ascent ({\bf GA}).}
Figure \ref{fig:ablation}a 
depicts the performance variation of two baseline algorithms, {\bf COMs}, and {\bf GA}, with and without our \ourmethod~regularizer. It is observed that initially these baselines outperform their \ourmethod-enhanced counterparts, but as the number of optimization steps increases, their performance starts to lag behind. This suggests that \ourmethod~will become increasingly beneficial for improving the baseline performance in the latter stages of the optimization process.

{\bf Choosing the sensitivity threshold \(\alpha\) -- see Definition~\ref{def:1}}. Figure \ref{fig:ablation}b 
visualizes how the performance of baselines regularized with \ourmethod~ is influenced by varying the value of \(\alpha\). The results indicate that using either an excessively low or high value for \(\alpha\) will impact the performance negatively. In all cases, the results suggest that a universal value of 0.1 for \(\alpha\) tend to generate consistent and effective performance across all tasks.

{\bf Choosing the no. $m$ of perturbation samples.}
Figure \ref{fig:ablation}c 
visualizes how the performance of baselines regularized with \ourmethod~ is influenced by varying the number of perturbation samples drawn from \(\mathbb{N}(\omega_\mu\mathbf{1}, \omega_\sigma^2\mathbf{I})\). It is observed that with more perturbation samples, \ourmethod~-regularized baseline achieves higher performance gain but also incurs more compute expense. As the performance gain beyond \(m = 100\) is marginal, we choose \(m = 100\) in all experiments.

{\bf Choosing value ranges for \(\omega\).}
As we mentioned previously in Section~\ref{sec:trustability}, large values for $\omega$ can make the sensitivity measure vacuous because by definition, sensitivity characterizes changes under slight perturbation of the model weight. To find this appropriate range, we plot the performance of the \textbf{GA} and \textbf{COMS} baselines regularized with \ourmethod~with respect to a set of potential ranges \(([\omega_{\mu_l},\omega_{\mu_u}],[\omega_{\sigma_l},\omega_{\sigma_u}])\) for \(\omega_\mu\) and \(\omega_\sigma\) in Figure \ref{fig:ablation}d. 
The results indicate that using an excessively low or high range of values will impact the performance negatively. Overall, the empirical results suggest that $[-10^{-3}, 10^{-3}]$ and $[10^{-5}, 10^{-2}]$ are best value ranges for $\omega_\mu$  for $\omega_\sigma$, respectively.  Additionally, we report the final performance of \ourmethod~with huge bound for $\omega$ in Appendix \ref{app:large_omega_bound}.

\begin{figure*}[t]
     \centering
     \begin{minipage}{0.39\linewidth} 
         \centering
         \includegraphics[width=0.85\textwidth]{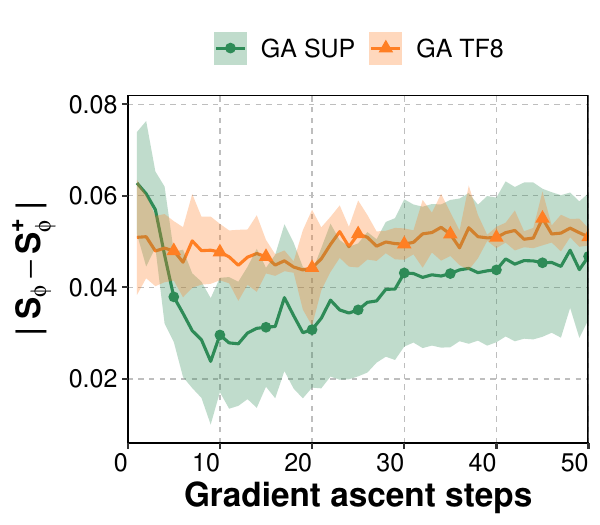}
         \caption{The mean and standard deviation of \(|S_\phi-S_\phi^+|\) across data batches during 50 epochs of \textbf{GA} on \textbf{Superconductor} and \textbf{TF-BIND-8}.}
         \label{fig:tightness}
     \end{minipage}
     \hfill
     \begin{minipage}{0.59\linewidth} 
        \centering
        \vspace{0.4cm}
         \includegraphics[width=\textwidth]{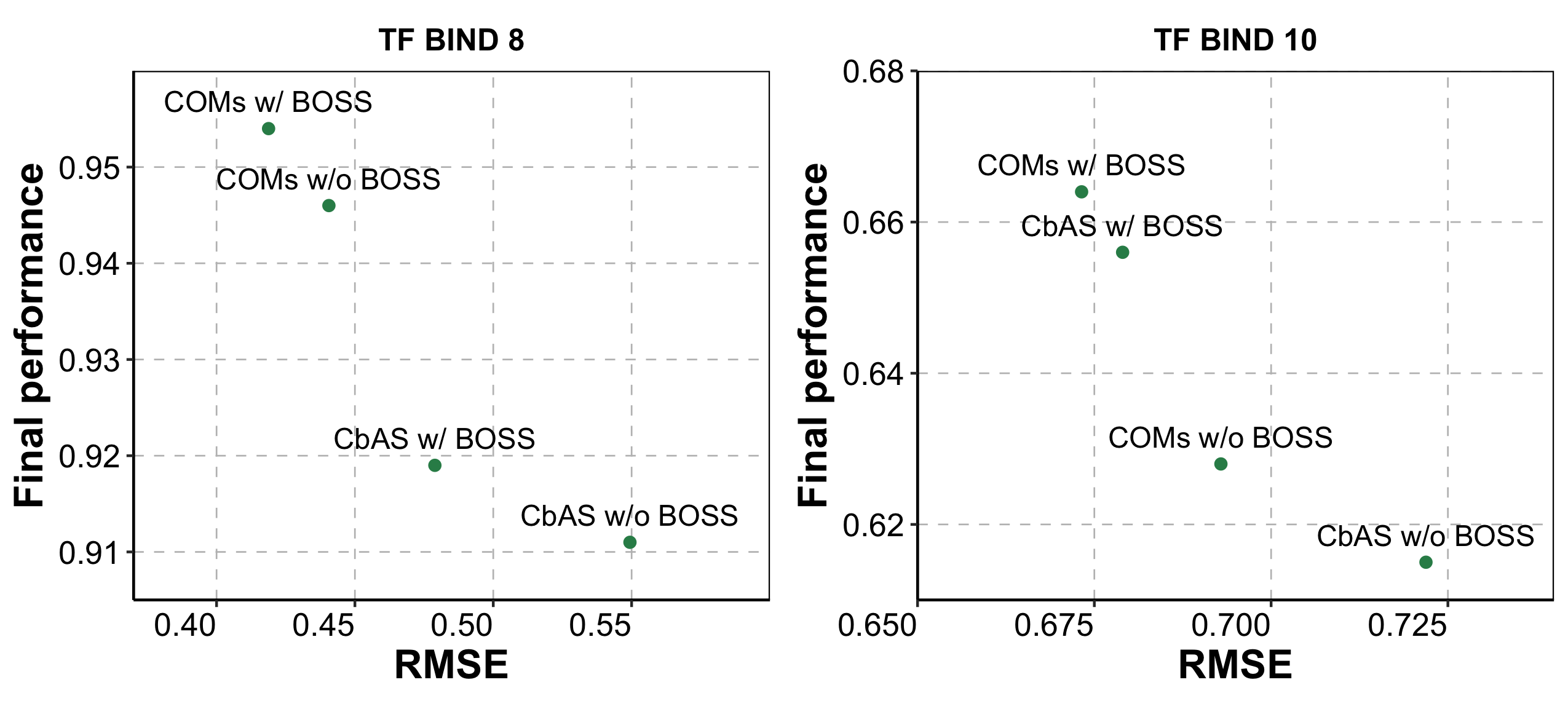}
        \caption{Improvement in terms of RMSE and correlation between the final performance with RMSE of \textbf{CBAS}, \textbf{COMs} on unseen data after being conditioned with \ourmethod~ on \textbf{TF-BIND-8} and \textbf{TF-BIND-10}.}
        \label{fig:correlation}
     \end{minipage}
\end{figure*}

{\bf Tightness of Lemma \ref{lem:2}.} Technically, Lemma~\ref{lem:2} can be made tighter by replacing $(A(\phi,\gamma)/\alpha)^2$ with $(A(\phi,\gamma)/\alpha)^n$ for $n > 2$ on the RHS of Eq.~\eqref{eq:12a}. This is however not necessary as our empirical studies suggest that even with $n = 2$, the gap between the $S_\phi$ and $S_\phi^+$ is already sufficiently small. This is detailed in Fig.~\ref{fig:tightness} below.
The plotted experiment results essentially reveal that the biggest difference between $S_\phi$ and $S_\phi^+$ is about 7.5\%, while the average is about 5\%. This establishes the tightness of Lemma \ref{lem:2}. 

\textbf{Validating the localized conditioning effect with relevant metrics.}
The localized conditioning effect in the offline optimizer arises because the offline data is not representative of the entire data space, causing the learned optimizer to overspecialize to the offline data.

This localized conditioning effect can be validated by examining the predictive performance (instead of the final optimization performance) of prior work's conditioned surrogate, both with and without our developed regularizer, on unseen input. An example of such validation is provided in Figure \ref{fig:additional_exp}b, which demonstrates that the root-mean-squared-error (RMSE) of \textbf{CBAS} on unseen data improves after being further conditioned with our proposed \ourmethod~ framework. This improvement is observed on both the \textbf{TF-BIND-8} and \textbf{TF-BIND-10} tasks.

Additionally, we conducted an experiment to assess the improvement in terms of RMSE and the correlation between the final performance and RMSE of \textbf{CBAS} and \textbf{COMs} on unseen data after being conditioned with \ourmethod~on \textbf{TF-BIND-8} and \textbf{TF-BIND-10}. The results of this experiment are presented in Figure \ref{fig:correlation}. Figure \ref{fig:correlation} conclusively illustrates that: (1) the localized conditioning effect exists (validated via RMSE); (2) localized conditioning can be significantly mitigated with our regularizer (the surrogate conditioned with our regularizer achieves better RMSE); and (3) there is a correlation between validation and the final metric (the surrogate with lower RMSE also achieves better optimization metrics).

\textbf{Computational Complexity}.
For clarity, we would like to detail the computational complexity of the \ourmethod~algorithm (i.e., Algorithm \ref{alg:pseudo-code}) below in terms of the number of parameters of the surrogate model $|\phi|$, the number of parameters of the neural approximation $|\Phi|$ (see Eq. (8)-(9)), the number of perturbation samples $m$, the number of training epochs $e$ needed to generate $\Phi$ in line 8, and the number of iterations $\tau$ in line 1 of Algorithm \ref{alg:pseudo-code}. Below is the per-line complexity breakdown that occurs within an outer loop over $\tau$ iterations in line 2:

Line 3: $O(m)$; Line 4: $O(|\phi|)$; Lines 5-6: $O(m|\phi|)$; Line 8: $O(me|\Phi|)$, where e is the number of epochs to train $\Phi$; Lines 10-11: $O(2m)$; Line 17: $O(m|\phi|)$; Line 18: $O(3|\phi|)$

Given that the above per-line complexity breakdown, the total computational complexity of Algorithm \ref{alg:pseudo-code} is

$O(\tau * (3m + 4|\phi| + 2m|\phi| + me|\Phi|)) = O(\tau * m * (|\phi| + |\Phi|))$

This indicates that the complexity is linear in $m$. Therefore, a linear increase in $m$ (e.g. $\Delta m$) will result in a corresponding linear increase in computational overhead (e.g. $O(\Delta m * \tau * (|\phi| + |\Phi|))$).

\section{Conclusion}
\label{sec:conclude}
This paper formalized the concept of model sensitivity in offline optimization, which inspires a new sensitivity-informed regularization that works synergistically with numerous existing approaches to boost their performance. As such, our contribution stands as an essential addition to the existing body of research in this field, providing a versatile and effective performance booster for a wide range of offline optimizers. This is extensively demonstrated on a diverse task benchmark. In addition, we believe the developed principles can also be adapted to related disciplines such as safe Bayesian optimization (BO) or safe reinforcement learning (RL) in online interactive learning scenarios, which are potential follow-up of our current work.

\section*{Impact Statement}
\label{sec:impact}
This research focuses on developing an effective regularizer for a wide range of existing offline optimization algorithms that help improve their performance. The mathematical approaches and insights developed in this paper will benefit various science and engineering applications, such as optimizing the design of hardware, materials, and molecules which require optimizing a black-box function using only its prior (offline) experimental data. Our experimental work uses publicly available datasets to evaluate the performance of our algorithms and we foresee no adverse ethical or societal consequences stemming from this research.

\section*{Acknowledgement}
This work was funded by Vingroup Joint Stock Company (Vingroup JSC),Vingroup, and supported by Vingroup Innovation Foundation (VINIF) under project code VINIF.2021.DA00128.

\nocite{langley00}

\bibliography{main}
\bibliographystyle{icml2024}


\newpage
\onecolumn
\appendix
\section{Derivation of Lemma~\ref{lem:1}}
\label{app:a}
The derivation of Lemma~\ref{lem:1} goes as follows. First, note that:
\begin{eqnarray}
\left|\mathbb{E}_{\mathbf{x} \sim \mathfrak{D}}\Big[g(\mathbf{x}; \phi + \gamma)\Big] \ -\  \mathbb{E}_{\mathbf{x} \sim \mathfrak{D}}\Big[g(\mathbf{x}; \phi)\Big]\right| &\leq& \mathbb{E}_{\mathbf{x} \sim \mathfrak{D}}\Big|g(\mathbf{x}; \phi + \gamma) - g(\mathbf{x}; \phi)\Big| \\
&\leq& \max_{\mathbf{x} \in \mathfrak{D}} \Big|g(\mathbf{x}; \phi + \gamma) - g(\mathbf{x}; \phi)\Big| \ .\label{eq:a1}
\end{eqnarray}
This implies the probability that the LHS is larger than $\alpha$ is smaller than the probability that the RHS is larger than $\alpha$. Hence,
\begin{eqnarray}
1 - \delta \ \ \leq\ \ \mathcal{S}_\phi(\alpha, \omega) &\leq& \underset{\gamma \sim \mathbb{N}(\omega_\mu, \omega_\sigma^2\mathbf{I})}{\mathrm{Pr}}\Bigg\{\max_{\mathbf{x} \in \mathfrak{D}}\ \Big|g(\mathbf{x}; \phi + \gamma) - g(\mathbf{x}; \phi)\Big| \ \geq\ \alpha \Bigg\}  \ ,\label{eq:a2}
\end{eqnarray}
which implies with probability at least $1 - \delta$, there exists an input at which the prediction might change by more than $\alpha$ due to a slight perturbation. This establishes the first part of Lemma~\ref{lem:1}.

Now, let this input be $\mathbf{x}$. Then, with probability at least $1 - \delta$:
\begin{eqnarray}
\alpha \ \ \leq\ \ \big|g(\mathbf{x}; \phi + \gamma) - g(\mathbf{x}; \phi) \big| &\leq& \big|g(\mathbf{x}; \phi + \gamma) - g(\mathbf{x}'; \phi + \gamma) \big| \nonumber\\
&+& \big|g(\mathbf{x}'; \phi + \gamma) - g(\mathbf{x}'; \phi) \big| \ +\ \big|g(\mathbf{x}'; \phi) - g(\mathbf{x}; \phi) \big| \nonumber\\
&\leq& 2\mathfrak{L}_{\phi}\|\mathbf{x} - \mathbf{x}'\| \ +\ \big|g(\mathbf{x}'; \phi + \gamma) - g(\mathbf{x}'; \phi) \big|  \label{eq:a3}
\end{eqnarray}
is true for any $\mathbf{x}' \in \mathfrak{D}$. To see this, the first step in \eqref{eq:a3} above follows from the quadrilateral generalization of the triangle inequality and the second step follows from the definition of Lipschitz constant in Lemma~\ref{lem:1}. Now, if $\mathbf{x}'$ is within a $(\alpha / 4\mathfrak{L}_\phi)$-ball centered at $\mathbf{x}$, we have $\|\mathbf{x} - \mathbf{x}'\| \leq \alpha / (4\mathfrak{L}_\phi)$. 

Plugging this into \eqref{eq:a3} leads to:
\begin{eqnarray}
\alpha &\leq& 2\mathfrak{L}_{\phi}\|\mathbf{x} - \mathbf{x}'\| \ +\ \big|g(\mathbf{x}'; \phi + \gamma) - g(\mathbf{x}'; \phi) \big| \nonumber\\
&\leq& 2\mathfrak{L}_\phi \cdot \frac{\alpha}{4\mathfrak{L}_\phi} \ +\ \big|g(\mathbf{x}'; \phi + \gamma) - g(\mathbf{x}'; \phi) \big| \ =\ \frac{\alpha}{2} \ +\ \big|g(\mathbf{x}'; \phi + \gamma) - g(\mathbf{x}'; \phi) \big| \ ,
\end{eqnarray}
which implies $|g(\mathbf{x}'; \phi + \gamma) - g(\mathbf{x}'; \phi)| \geq \alpha - \alpha / 2 = \alpha / 2$. Thus, with probability $1 - \delta$, this is true for all $\mathbf{x}'$ in the $(\alpha / 4\mathfrak{L}_\phi)$-ball centered at $\mathbf{x}$. The second part of Lemma~\ref{lem:1} has been derived.

\section{Derivation of Lemma~\ref{lem:2}}
\label{app:c}
The derivation of Lemma~\ref{lem:2} goes as follows. First, note that:
\begin{eqnarray}
\mathcal{S}_\phi\left(\alpha, \omega\right) &\triangleq& \underset{\gamma \sim \mathbb{N}(\omega_\mu, \omega_\sigma^2\mathbf{I})}{\mathrm{Pr}}\Bigg(\ \left|\mathbb{E}_{\mathbf{x} \sim \mathfrak{D}}\Big[g(\mathbf{x}; \phi + \gamma)\Big] \ -\  \mathbb{E}_{\mathbf{x} \sim \mathfrak{D}}\Big[g(\mathbf{x}; \phi)\Big]\right| \ \ \geq\ \ \alpha \ \Bigg)\\  & = & \underset{\gamma \sim \mathbb{N}(\omega_\mu, \omega_\sigma^2\mathbf{I})}{\mathbb{E}} \Bigg[\ \mathbb{I}\left(\left|\mathbb{E}_{\mathbf{x} \sim \mathfrak{D}}\Big[g(\mathbf{x}; \phi + \gamma)\Big] \ -\  \mathbb{E}_{\mathbf{x} \sim \mathfrak{D}}\Big[g(\mathbf{x}; \phi)\Big]\right| \ \ \geq\ \ \alpha \right) \ \Bigg] \ .
\end{eqnarray}

On the other hand, we have 
\begin{eqnarray}
    A &=& \mathbb{I}\left(\left|\mathbb{E}_{\mathbf{x} \sim \mathfrak{D}}\Big[g(\mathbf{x}; \phi + \gamma)\Big] \ -\  \mathbb{E}_{\mathbf{x} \sim \mathfrak{D}}\Big[g(\mathbf{x}; \phi)\Big]\right| \geq \alpha \right) \\
    & < & \frac{1}{\alpha^n}\cdot\left|\mathbb{E}_{\mathbf{x} \sim \mathfrak{D}}\Big[g(\mathbf{x}; \phi + \gamma)\Big] \ -\  \mathbb{E}_{\mathbf{x} \sim \mathfrak{D}}\Big[g(\mathbf{x}; \phi)\Big]\right|^n \ ,
\end{eqnarray}
which is true for any \(\gamma \sim \mathbb{N}(\omega_\mu, \omega_\sigma^2\mathbf{I})\) and \(\alpha > 0\). Thus, choosing \(n = 2\) results in a differentiable (with respect to $\phi$) function that upper bounds the regularizer as follows:
\begin{eqnarray}
    \mathcal{S}_\phi\left(\alpha, \omega\right) & < & \frac{1}{\alpha^2} \cdot \underset{\gamma \sim \mathbb{N}(\omega_\mu, \omega_\sigma^2\mathbf{I})}{\mathbb{E}} \Bigg[\ \left|\mathbb{E}_{\mathbf{x} \sim \mathfrak{D}}\Big[g(\mathbf{x}; \phi + \gamma)\Big] \ -\  \mathbb{E}_{\mathbf{x} \sim \mathfrak{D}}\Big[g(\mathbf{x}; \phi)\Big]\right|^2\ \Bigg] \ .
\end{eqnarray}
However, when the difference between the surrogate model's prediction and the perturbed model's prediction is considerably larger than alpha, the upper bound will be significantly higher than the original value, leading to reduced reliability of the upper bound. To address this issue, we employ the minimum function to set a threshold on the disparity between the upper bound regularizer and the original regularizer. By using the min function, we aim to reduce the difference between the upper bound regularizer and the original one, without affecting the computation of gradients since the min function is differentiable. Hence,
\begin{eqnarray}
\hspace{-19mm}\mathcal{S}_\phi\left(\alpha, \omega\right) & < & \underset{\gamma \sim \mathbb{N}(\omega_\mu, \omega_\sigma^2\mathbf{I})}{\mathbb{E}} \left[\ \operatorname*{min} \left( 1, \frac{1}{\alpha^2} \cdot \left|\mathbb{E}_{\mathbf{x} \sim \mathfrak{D}}\Big[g(\mathbf{x}; \phi + \gamma)\Big] \ -\  \mathbb{E}_{\mathbf{x} \sim \mathfrak{D}}\Big[g(\mathbf{x}; \phi)\Big]\right|^2 \right) \ \right] \ \ \triangleq\ \  \mathcal{S}^+_\phi(\alpha, \omega) \ .
\end{eqnarray}

\section{Alternative Formulation of \ourmethod~: \variantmethod~ via using Taylor Approximation in Eq.~\eqref{eq:8}}
\label{sec:SIRO-CDF}
Alternative to our main formulation of \ourmethod~ regularization, we can also apply Taylor approximation directly to the definition of $S_\phi(\alpha, \omega)$ as follows:
\begin{eqnarray}
\mathcal{S}_\phi\left(\alpha, \omega\right) &\triangleq& \underset{\gamma \sim \mathbb{N}(\omega_\mu, \omega_\sigma^2\mathbf{I})}{\mathrm{Pr}}\Bigg(\ \left|\mathbb{E}_{\mathbf{x} \sim \mathfrak{D}}\Big[g(\mathbf{x}; \phi + \gamma)\Big] \ -\  \mathbb{E}_{\mathbf{x} \sim \mathfrak{D}}\Big[g(\mathbf{x}; \phi)\Big]\right| \ \ \geq\ \ \alpha \ \Bigg) \ .\label{eq:c1}
\end{eqnarray}
Instead of using the neural re-parameterization (as detailed in the main text), we could approximate $h(\phi + \gamma) \triangleq \mathbb{E}_{\mathbf{x} \sim \mathfrak{D}}\Big[g(\mathbf{x}; \phi + \gamma)\Big]$ with a first-order Taylor expansion around $\phi$. That is,
\begin{eqnarray}
\mathbb{E}_{\mathbf{x} \sim \mathfrak{D}}\Big[g(\mathbf{x}; \phi + \gamma)\Big] \ -\  \mathbb{E}_{\mathbf{x} \sim \mathfrak{D}}\Big[g(\mathbf{x}; \phi)\Big]  &=& h(\phi + \gamma) - h(\phi) \ \ \simeq\ \ \nabla h(\phi)^\top \gamma \ . \label{eq:c2}
\end{eqnarray}
Plugging Eq.~\eqref{eq:c2} into Eq.~\eqref{eq:c1} leads to
\begin{eqnarray}
\mathcal{S}_\phi\left(\alpha, \omega\right) &=& \underset{\gamma \sim \mathbb{N}(\omega_\mu, \omega_\sigma^2\mathbf{I})}{\mathrm{Pr}}\Bigg(\ \left|\nabla h(\phi)^\top \gamma\right| \ \ \geq\ \ \alpha \ \Bigg) \ .
\end{eqnarray}
Since $\gamma$ follows the Gaussian distribution $\mathbb{N}(\omega_\mu, \omega_\sigma^2\mathbf{I})$, we have $\nabla h(\phi)^\top \gamma$ follows the univariate Gaussian distribution $\mathbb{N}\left(\ \omega_\mu^\top \nabla h(\phi), \ \nabla h(\phi)^\top  \omega_\sigma^2\mathbf{I} \ \nabla h(\phi) \ \right) = \mathbb{N} (\mu_z,\sigma^2_z)$ where $\mu_z \triangleq \omega_\mu^\top \nabla h(\phi)$ and $\sigma_z^2 \triangleq \omega_\sigma^2\nabla h(\phi)^\top\nabla h(\phi)$ as a direct consequence. Now, let $z \triangleq \nabla h(\phi)^\top\gamma$,
\begin{eqnarray}
\mathcal{S}_\phi\left(\alpha, \omega\right) &\triangleq& \underset{z \sim \mathbb{N} (\mu_z,\sigma^2_z)}{\mathrm{Pr}}\Big(\ \left|z\right| \ \ \geq\ \ \alpha \ \Big) \nonumber \\
&=& \underset{z \sim \mathbb{N} (\mu_z,\sigma^2_z)}{\mathrm{Pr}}\Big(\ z \ \ \leq\ \ -\alpha \ \Big) \ +\  \underset{z \sim \mathbb{N} (\mu_z,\sigma^2_z)}{\mathrm{Pr}}\Big(\ z \ \ \geq\ \ \alpha \ \Big) \nonumber \\
&=& \underset{z \sim \mathbb{N} (\mu_z,\sigma^2_z)}{\mathrm{Pr}}\Big(\ z \ \ \leq\ \ -\alpha \ \Big) \ +\ 1 \ - \underset{z \sim \mathbb{N} (\mu_z,\sigma^2_z)}{\mathrm{Pr}}\Big(\ z \ \ \leq\ \ \alpha \ \Big) \nonumber \\
&=& 1 \ + \ \mathrm{F}_z(-\alpha) \ - \ \mathrm{F}_z(\alpha) \ ,
\label{eq:30}
\end{eqnarray}
where $\mathrm{F}_z(.)$ is the cumulative distribution function (CDF) of the real-valued random variable $z \sim \mathbb{N} (\mu_z,\sigma^2_z)$. As the direct approximation in Eq.~\eqref{eq:30} is already differentiable, we can bypass the use of a differentiable upper-bound in Lemma~\ref{lem:2} and the neural re-parameterization in Eq.~\eqref{eq:9} to optimize $S_\phi(\alpha, \omega)$ via its (differentiable) proxy in Eq.~\eqref{eq:30}. Despite its simplified formulation, its empirical improvement over the baseline performance is less pronouncing than the improvement achieved by our main method~\ourmethod. This is reported in Table~\ref{tb:siro_cdf} below where the alternative approach \variantmethod~ is compared with our main method \ourmethod~ on 3 baselines: \textbf{GA}, \textbf{REINFORCE}, and \textbf{COMs} over $6$ benchmark tasks. 

In particular, the results in Table \ref{tb:siro_cdf} show that \variantmethod~ also improves well over the baseline performance in the majority of cases. For example, \variantmethod~ consistently enhances the baseline performance in $13/18$ instances, with a maximum and average improvement of $3.1\%$ and $1.25\%$, respectively.  However, our main method \ourmethod~ still achieves better and more consistent improvement than \variantmethod~. Using \ourmethod, we observe no task instances with degradation over baseline performance. Furthermore, \ourmethod~ also outperforms \variantmethod~ in $12/18$ cases with a maximum and average difference of $29.3\%$ and $3.78\%$, respectively. In contrast, \variantmethod~ only improves slightly over \ourmethod~ in $6/18$ cases with a small (average) margin of $0.5\%$.



\begin{table*}[h]
    \caption{Percentage of performance improvement achieved by \ourmethod~across all tasks and baselines at the \textbf{$100$-th percentile} level. \textbf{P} denotes the normalized performance while \textbf{G} denotes \ourmethod's and \variantmethod's percentage of gain over the baseline performance.}
    \centering
    \setlength\tabcolsep{3pt} 
    \resizebox{\textwidth}{!}{
        \begin{tabular}{ll|ll|ll|ll|ll|ll|ll} 
         \toprule
           & & \multicolumn{6}{c|}{\textbf{Continuous Tasks}} & \multicolumn{6}{c}{\textbf{Discrete Tasks}} 
           \\\cmidrule(lr){3-8}\cmidrule(lr){9-14}
           & & \multicolumn{2}{c}{\textbf{Ant Morphology}} & \multicolumn{2}{c}{\textbf{D'Kitty Morphology}} & \multicolumn{2}{c|}{\textbf{Superconductor}} 
          & \multicolumn{2}{c}{\textbf{TF Bind 8}} & \multicolumn{2}{c}{\textbf{TF Bind 10}} & \multicolumn{2}{c}{\textbf{ChEMBL}}
          
          \\ \cmidrule(lr){3-4}\cmidrule(lr){5-6} \cmidrule(lr){7-8}\cmidrule(lr){9-10}\cmidrule(lr){11-12}\cmidrule(lr){13-14}

          \multicolumn{2}{c|}{\textbf{Algorithms}}& \multicolumn{1}{c}{\textbf{P}} & \multicolumn{1}{c}{\textbf{G}}
           & \multicolumn{1}{c}{\textbf{P}} & \multicolumn{1}{c}{\textbf{G}}
           & \multicolumn{1}{c}{\textbf{P}} & \multicolumn{1}{c|}{\textbf{G}}
           & \multicolumn{1}{c}{\textbf{P}} & \multicolumn{1}{c}{\textbf{G}}
           & \multicolumn{1}{c}{\textbf{P}} & \multicolumn{1}{c}{\textbf{G}}
           & \multicolumn{1}{c}{\textbf{P}} & \multicolumn{1}{c}{\textbf{G}}
          \\ \midrule

         \multirow{4}{*}{\textbf{GA}} & Base 
         & 0.299 ± 0.037 & 
         & 0.871 ± 0.012 & 
         & 0.506 ± 0.008 & 
         & 0.980 ± 0.015 & 
         & 0.647 ± 0.029 & 
         & 0.640 ± 0.010 & \\
         & \ourmethod
         & 0.314 ± 0.034 & \color{blue}+1.5\% 
         & 0.883 ± 0.012 & \color{blue}+1.2\% 
         & 0.515 ± 0.014 & \color{blue}+0.9\% 
         & 0.986 ± 0.007 & \color{blue}+0.6\% 
         & 0.658 ± 0.072 & \color{blue}+1.1\% 
         & 0.646 ± 0.002 & \color{blue}+0.6\% \\
         & \variantmethod
         &  0.323 ± 0.051 & \color{blue}+2.4\% 
         &  0.881 ± 0.013 & \color{blue}+1.0\% 
         &  0.527 ± 0.012 & \color{blue}+1.1\% 
         &  0.985 ± 0.012 & \color{blue}+0.5\% 
         &  0.645 ± 0.029 & \color{red}-0.2\% 
         &  0.640 ± 0.009 & \color{blue}+0.0\%
         \\ \midrule

         \multirow{3}{*}{\textbf{REINFORCE}} & Base 
         & 0.253 ± 0.047 & 
         & 0.674 ± 0.138 & 
         & 0.481 ± 0.015 & 
         & 0.929 ± 0.031 & 
         & 0.664 ± 0.061 & 
         & 0.634 ± 0.002 & \\
         & \ourmethod
         & 0.278 ± 0.014 & \color{blue}+2.5\% 
         & 0.732 ± 0.003 & \color{blue}+5.8\% 
         & 0.483 ± 0.007 & \color{blue}+0.2\% 
         & 0.943 ± 0.029 & \color{blue}+1.4\% 
         & 0.964 ± 0.096 & \color{blue}+30.0\%
         & 0.639 ± 0.010 & \color{blue}+0.5\% \\
         & \variantmethod 
         &  0.284 ± 0.045 & \color{blue}+3.1\% 
         &  0.696 ± 0.083 & \color{blue}+2.2\% 
         &  0.443 ± 0.012 & \color{red}-3.8\% 
         &  0.938 ± 0.047 & \color{blue}+0.9\% 
         &  0.671 ± 0.037 & \color{blue}+0.7\% 
         &  0.640 ± 0.015 & \color{blue}+0.6\% 
         \\ \midrule

         \multirow{3}{*}{\textbf{COMs}} & Base 
         & 0.896 ± 0.024 & 
         & 0.937 ± 0.012 & 
         & 0.483 ± 0.026 & 
         & 0.946 ± 0.035 & 
         & 0.628 ± 0.044 & 
         & 0.633 ± 0.000 & \\
        & \ourmethod  
        & 0.918 ± 0.025 & \color{blue}+2.2\% 
        & 0.942 ± 0.011 & \color{blue}+0.5\% 
        & 0.486 ± 0.030 & \color{blue}+0.3\% 
        & 0.954 ± 0.020 & \color{blue}+0.8\% 
        & 0.664 ± 0.038 & \color{blue}+3.6\%
        & 0.638 ± 0.009 & \color{blue}+0.5\% \\
        & \variantmethod  
        &  0.900 ± 0.024 & \color{blue}+0.4\% 
        &  0.944 ± 0.010 & \color{blue}+0.7\% 
        &  0.496 ± 0.019 & \color{blue}+1.3\% 
        &  0.943 ± 0.025 & \color{red}-0.3\% 
        &  0.641 ± 0.041 & \color{blue}+1.3\% 
        &  0.633 ± 0.000 & \color{blue}+0.0\% 
        \\\bottomrule
        \end{tabular}
    }
    \label{tb:siro_cdf}
\end{table*}

\section{Additional Experiment Results}
\label{app:b}
\begin{figure}[bt]
     \centering
     \begin{minipage}{0.48\linewidth} 
        \centering
         \includegraphics[width=\textwidth]{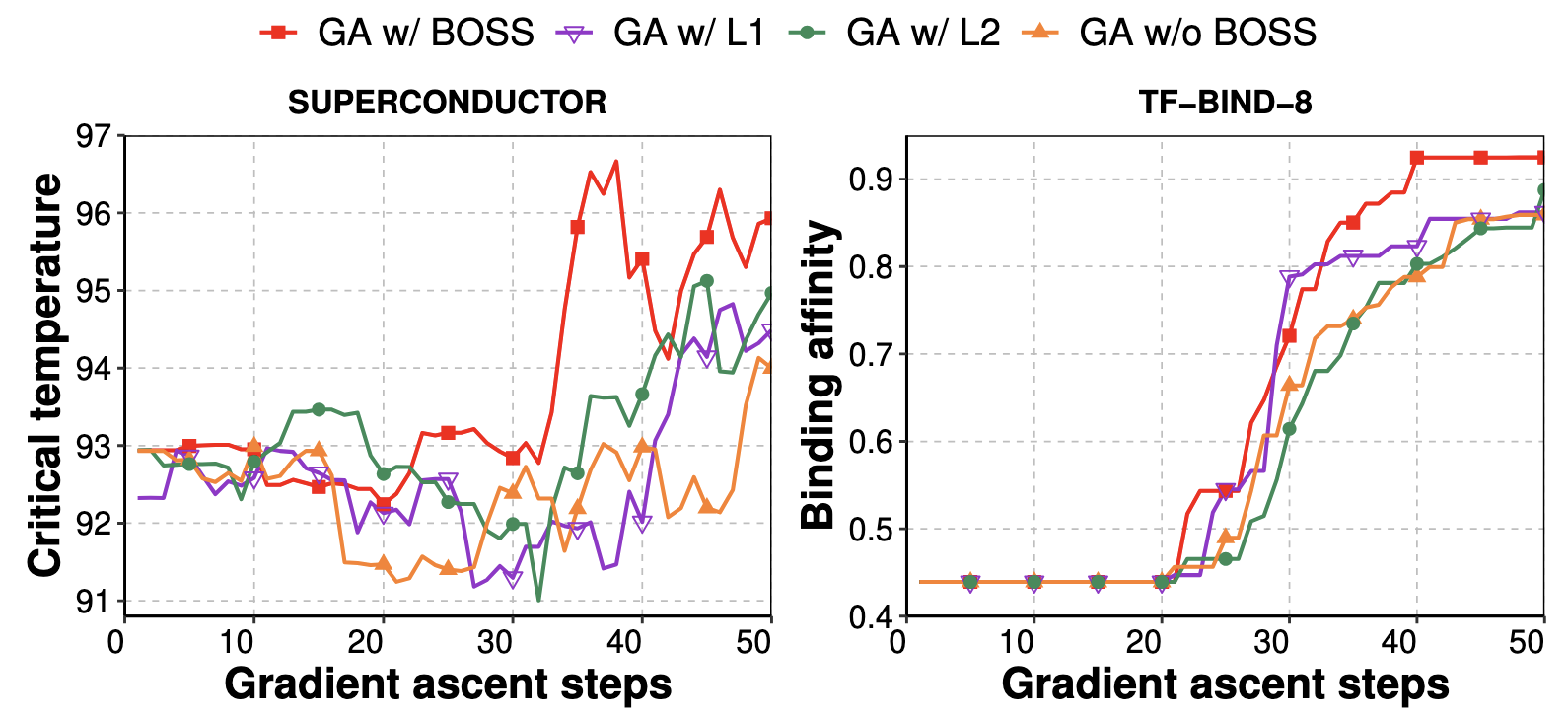}
         (a)
         \label{fig:L1_L2}
     \end{minipage}
     \begin{minipage}{0.48\linewidth} 
        \centering
         \includegraphics[width=\textwidth]{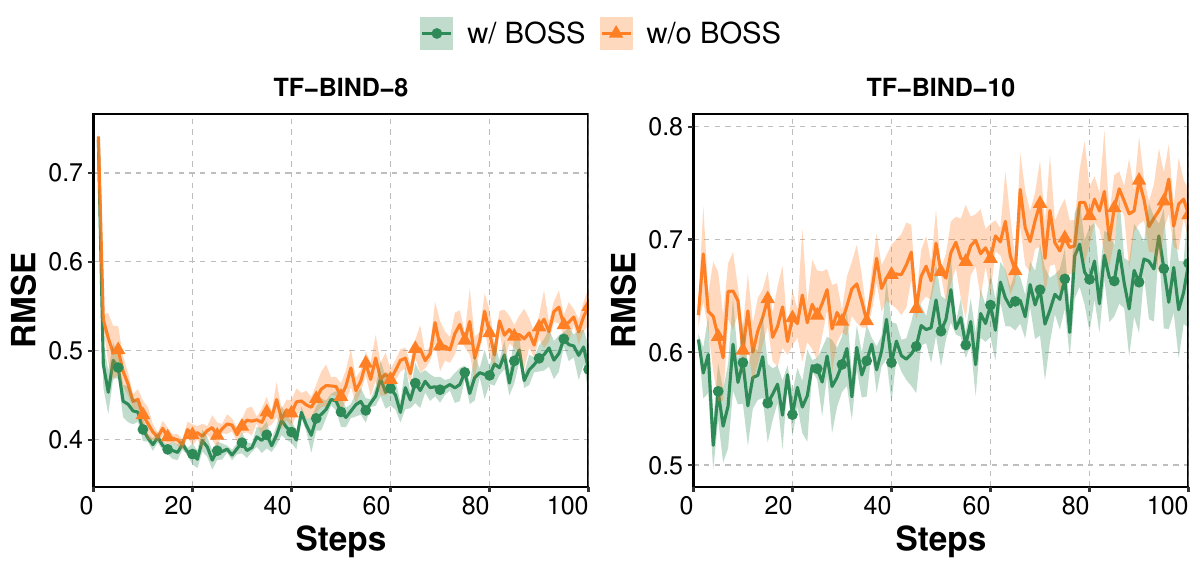}
         (b)
        \label{fig:RMSE_cbas}
     \end{minipage}
     
     \caption{Additional experiments: (a) Comparison \ourmethod ~with \(L1\) and \(L2\) regularization; (b) Root-mean-square-error (RMSE) of surrogate model trained with and without \ourmethod~ in simulated out-of-distribution regime.}
     \label{fig:additional_exp}
\end{figure}

\begin{figure}[bt]
     \centering
     \begin{minipage}{0.48\linewidth} 
        \centering
         \includegraphics[width=\textwidth]{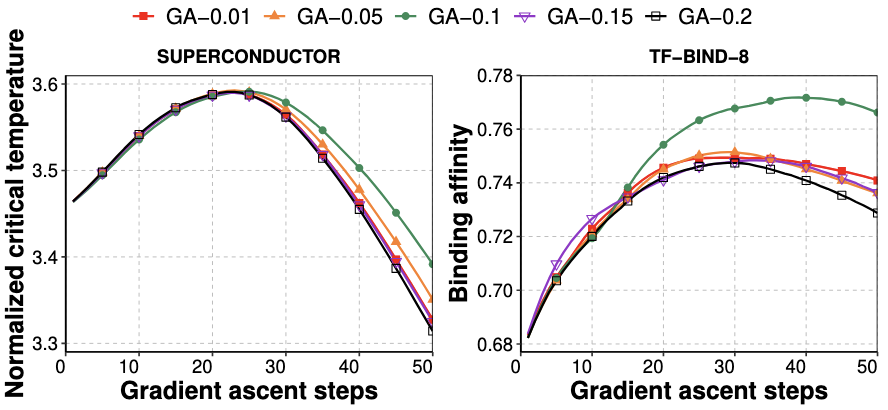}
         \\(a)
         \label{fig:coms_oracle_change_alpha}
     \end{minipage}
     \hfill
     \begin{minipage}{0.48\linewidth} 
        \centering
         \includegraphics[width=\textwidth]{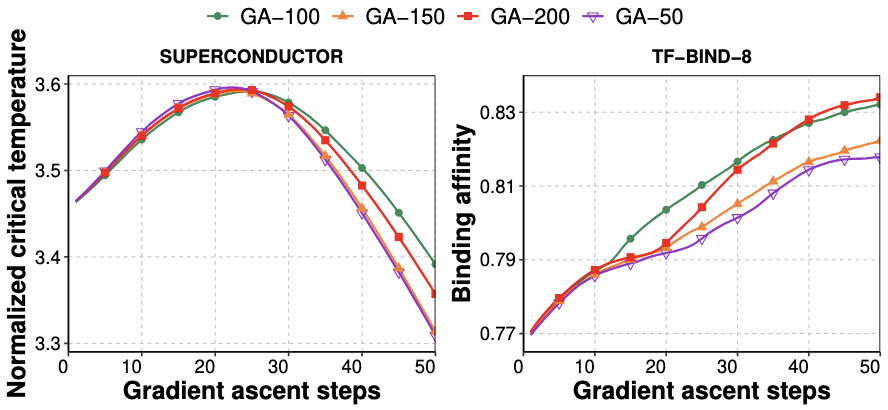}
         \\(b)
         \label{fig:coms_oracle_change_m}
    \end{minipage}
     \begin{minipage}{0.48\linewidth} 
        \centering
         \includegraphics[width=\textwidth]{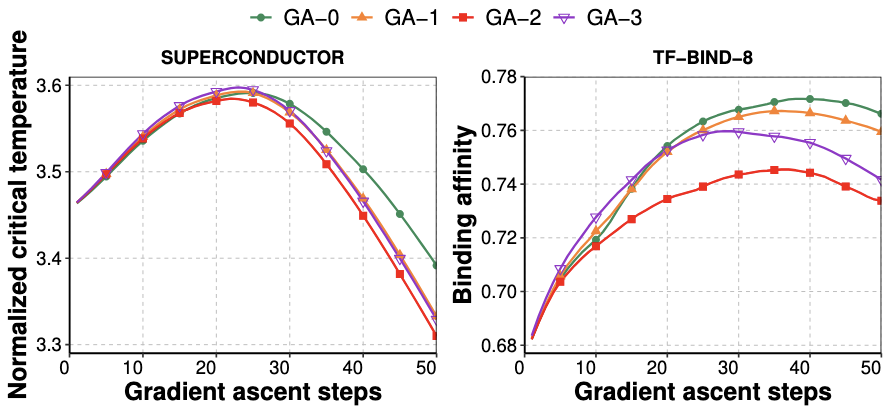}
         \\(c)
         \label{fig:coms_oracle_change_omega_bound}
    \end{minipage}
     \caption{Plots of performance (evaluated by \textbf{COMs} pseudo-oracle) variation of \textbf{GA} (regularized by \ourmethod) to changes in (a) values of the sensitivity threshold $\alpha$; (b) changes in the no. of perturbation samples $m$; and (c) changes in value ranges of bound \(([\omega_{\mu_l},\omega_{\mu_u}],[\omega_{\sigma_l},\omega_{\sigma_u}])\) for parameter $\omega$ of the perturbation distribution. These are \(([-10^{-3},10^{-3}],[10^{-5},10^{-2}])\), \( ([-10^{-3},10^{-3}],[10^{-6},10^{-3}])\), \( ([-10^{-2},10^{-2}],[10^{-5},10^{-2}])\), and \( ([-10^{-2},10^{-2}],[10^{-6},10^{-3}])\) which are indexed with 0, 1, 2, 3 in this figure.}\vspace{-6mm}
    \label{fig:coms_oracle}
\end{figure}

\subsection{Comparison with Other Regularization Methods}
In the realm of machine learning, traditional \(L1\) and \(L2\)-norm regularization techniques are known to prevent overfitting during model training. To evaluate their effectiveness, we carried out a compact experiment to juxtapose these methods with our regularization approach. This comparison involves examining the objective values of 128 solution designs through 50 steps of Gradient Ascent in two distinct tasks: \textbf{SUPERCONDUCTOR} and \textbf{TF-BIND-8}. Figure \ref{fig:additional_exp}a 
indicates that while \(L1\) and \(L2\) norms improve upon the baseline solution, they do not surpass the performance of our regularization technique. This is expected since \ourmethod ~is specifically designed to condition the output behavior of the model against adversarial perturbation while \(L1\) and \(L2\) only generically penalize models with high complexity, measured by the \(L1\) and \(L2\) norms of their parameters.

\subsection{Precision of Surrogate Prediction using \ourmethod}
Our regularization technique is designed to develop a more resilient surrogate model capable of handling minor disturbances and delivering accurate predictions in out-of-distribution regimes. To assess the precision of predictions using \ourmethod ~versus the original baseline, we carried out a small-scale experiment. In this experiment, we trained surrogate models with and without the \ourmethod~regularizer on the offline dataset and subsequently computed their root-mean-square-error (RMSE) on a test dataset to assess their prediction precision on a simulated out-of-distribution regime. This is based on the observation that the test dataset comprises samples with higher objective values than those of the offline dataset. The experimental results are illustrated in Figure \ref{fig:additional_exp}b which plots the mean and standard deviation of RMSE across multiple runs. The figures showed that the integration of \ourmethod~ into surrogate training resulted in a model that produces more accurate prediction than the model learned without using \ourmethod. This can be seen by observing a lower RMSE and reduced standard deviation in settings where the surrogate training was regularized with~\ourmethod. This observation is consistent across both the \textbf{TF-BIND-8} and \textbf{TF-BIND-10} tasks.

\subsection{Tuning Hyper-parameters of the Proposed Regularizer}
\label{app:b3}

To fine-tune the hyper-parameters of our regularizer, we can use the surrogate of the base method (which was trained on the offline data) as a pseudo-oracle, which helps evaluate design proposals generated by each specific configuration of the regularizer. For example, we can leverage the surrogate model of \textbf{COMs} as a pseudo-oracle to evaluate design proposals generated by the gradient ascent method \textbf{(GA)}.

To demonstrate this, we conducted three experiments that mirrored those in our ablation studies in Section \ref{sec:ablation} with the key difference being the use of a pseudo-oracle in place of the true oracle. The results are promising, showing that this technique can successfully discover the same optimal hyperparameters as those determined using the true oracle. Figure \ref{fig:coms_oracle}a shows that the value of sensitivity threshold \(\alpha = 0.1\) (GA-0.1) is the best, as previously demonstrated using true oracle in Figure \ref{fig:ablation}b. 
Likewise, Figure \ref{fig:coms_oracle}b 
shows that number of perturbation sample \(m=100\)
 as the corresponding regularized baseline GA-100 is the best among those in that plot. Figure \ref{fig:coms_oracle}c 
 shows the bounds on mean and variance of the perturbation distribution  \(([-10^{-3},10^{-3}],[10^{-5},10^{-2}])\)
 as the corresponding regularized baseline GA-0 is best. These are the same tuning results found by the oracle in our ablation studies in Figure \ref{fig:ablation}c and Figure \ref{fig:ablation}d. 
 Note that we only use the true oracle in the ablation studies (Section \ref{sec:ablation}) which is necessary to show the isolated effect of each of the components. Our other experiments do not use the true oracle for hyper-parameter tuning. It is also noted that the indexing of baseline in Figure \ref{fig:coms_oracle}a, \ref{fig:coms_oracle}b, and \ref{fig:coms_oracle}c 
 correspond to different indexing systems of the tuning parameter candidates.

\subsection{Comparison with Other Baselines}
\label{app:compare_ddom}
We also conducted additional experiments to compare the optimal performance achieved by \ourmethod~ with two recent algorithms: \textbf{DDOM} \citep{krishnamoorthy2023diffusion} and \textbf{tri\_mentoring} \citep{chen2023parallel}. The results reported in Table \ref{tb:boss_vs_ddom} indicate that \ourmethod~ also significantly outperforms these new baselines in almost all tasks, except for \textbf{D'Kitty Morphology} where our performance is slightly behind \textbf{tri\_mentoring}. We note that it is also possible to integrate the \ourmethod~regularizer to potentially improve the performance of \textbf{tri\_mentoring} similar to how it was done with \textbf{ICT}. 
\begin{table*}[h]
    \caption{Performance achieved by \ourmethod~, \textbf{DDOM}, and \textbf{tri\_mentoring} across all benchmark tasks at \textbf{$100$-th percentile}. We are not able to run the released code of \textbf{tri\_mentoring} on the \textbf{ChEMBL} task.}
    \centering
    \setlength\tabcolsep{3pt} 
    \resizebox{\textwidth}{!}{
        \begin{tabular}{c|c|c|c|c|c|c} 
         \toprule
           \multirow{2}{*}{\textbf{Algorithms}}& \multicolumn{3}{c|}{\textbf{Continuous Tasks}} & \multicolumn{3}{c}{\textbf{Discrete Tasks}}
           \\\cmidrule(lr){2-4}\cmidrule(lr){5-7}
           & \textbf{Ant Morphology} & \textbf{D'Kitty Morphology} & \textbf{Superconductor} 
          & \textbf{TF Bind 8} & \textbf{TF Bind 10} & \textbf{ChEMBL}
          

          \\ \midrule

        \textbf{DDOM} 
         & 0.920 ± 0.013  
         & 0.934 ± 0.007  
         & 0.481 ± 0.037  
         & 0.946 ± 0.022  
         & 0.668 ± 0.075  
         & 0.635 ± 0.005 \\ 
          \textbf{tri\_mentoring}
         & 0.932 ± 0.014  
         & \textbf{0.952 ± 0.011}  
         & 0.509 ± 0.012 
         & 0.927 ± 0.015  
         & 0.665 ± 0.011  
         &  N/A \\ 
          \textbf{\ourmethod}
         & \textbf{0.939 ± 0.013}
         & 0.943 ± 0.013  
         & \textbf{0.515 ± 0.014}  
         & \textbf{0.989 ± 0.007}  
         & \textbf{0.964 ± 0.096} 
         & \textbf{0.688 ± 0.000} 
        \\\bottomrule
        \end{tabular}
    }
    \label{tb:boss_vs_ddom}
\end{table*}

\subsection{Performance Evaluation at 75-th and 50-th Percentile Level}\label{app:75_50_percentile}

According to the reported results in Table \ref{tb:75th_results}, the no. of cases where there is a slight performance decrease is $20/66$. The maximum decrease across all such cases is $1.9\%$. This means in $46/66 = 69.7\%$ of the cases the performance is either preserved or improved. The performance also strictly increases in $24$ cases with a maximum and average improvement of $23.3\%$ and $2.104\%$, respectively. On the other hand, the average decrease is only $0.59\%$. Likewise, in Table \ref{tb:50th_results}, the observed average improvement is $2.233\%$ while the average decrease is only $0.952\%$. Following the protocol in~\cite{TrabuccoICML21} (COMS), the offline optimizer starts with $128$ initial points and performs a number of search steps to arrive at $128$ solution candidates. These solutions are sorted in increasing order such that the last point corresponds to the $100$-th percentile, the $3$-rd quarter point corresponds to the $75$-th percentile and the middle point corresponds to the $50$-th percentile. Consequently, the result at the $100$-th percentile is better than the results at the $75$-th and $50$-th percentile settings as expected. 

\begin{table}[tbh]
\caption{Percentage of performance improvement achieved by \ourmethod~across all tasks and baselines at the \textbf{$75$-th percentile} level. \textbf{P} denotes the normalized performance while \textbf{G} denotes \ourmethod's percentage of gain over the baseline performance.}
    \centering
    \setlength\tabcolsep{3pt} 
    \resizebox{\textwidth}{!}{
        \begin{tabular}{ll|ll|ll|ll|ll|ll|ll} 
         \toprule
           & & \multicolumn{6}{c|}{\textbf{Continuous Tasks}} & \multicolumn{6}{c}{\textbf{Discrete Tasks}} 
           \\\cmidrule(lr){3-8}\cmidrule(lr){9-14}
           & & \multicolumn{2}{c}{\textbf{Ant Morphology}} & \multicolumn{2}{c}{\textbf{D'Kitty Morphology}} & \multicolumn{2}{c|}{\textbf{Superconductor}} 
          & \multicolumn{2}{c}{\textbf{TF Bind 8}} & \multicolumn{2}{c}{\textbf{TF Bind 10}} & \multicolumn{2}{c}{\textbf{ChEMBL}}
          
          \\ \cmidrule(lr){3-4}\cmidrule(lr){5-6} \cmidrule(lr){7-8}\cmidrule(lr){9-10}\cmidrule(lr){11-12}\cmidrule(lr){13-14}

          \multicolumn{2}{c|}{\textbf{Algorithms}}& \multicolumn{1}{c}{\textbf{P}} & \multicolumn{1}{c}{\textbf{G}}
           & \multicolumn{1}{c}{\textbf{P}} & \multicolumn{1}{c}{\textbf{G}}
           & \multicolumn{1}{c}{\textbf{P}} & \multicolumn{1}{c|}{\textbf{G}}
           & \multicolumn{1}{c}{\textbf{P}} & \multicolumn{1}{c}{\textbf{G}}
           & \multicolumn{1}{c}{\textbf{P}} & \multicolumn{1}{c}{\textbf{G}}
           & \multicolumn{1}{c}{\textbf{P}} & \multicolumn{1}{c}{\textbf{G}}
          \\ \midrule 
         \(\mathfrak{D}\)(\textbf{best}) & & 0.565 & & 0.884 & & 0.400 & & 0.439 & & 0.467 & & 0.605
         \\ \midrule
        
        \multirow{2}{*}{\textbf{CbAS}}  & Base
        & 0.523 ± 0.037 &
        & 0.797 ± 0.009 &
        & 0.195 ± 0.014 & 
        & 0.534 ± 0.015 &
        & 0.496 ± 0.009 & 
        & 0.633 ± 0.000 &\\
        & \ourmethod 
        & 0.522 ± 0.057 & \color{red}-0.1\% 
        & 0.794 ± 0.007 & \color{red}-0.3\% 
        & 0.209 ± 0.012 & \color{blue}+1.4\% 
        & 0.531 ± 0.033 & \color{red}-0.3\% 
        & 0.505 ± 0.011 & \color{blue}+0.9\%
        & 0.633 ± 0.000 & \color{blue}+0.0\%
        \\ \midrule
        
        \multirow{2}{*}{\textbf{BO-qEI}} & Base 
        & 0.607 ± 0.000  & 
        & 0.884 ± 0.000  & 
        & 0.306 ± 0.020  & 
        & 0.439 ± 0.000  & 
        & 0.502 ± 0.007  & 
        & 0.629 ± 0.005 & \\
        & \ourmethod
        & 0.607 ± 0.000 & \color{blue}+0.0\% 
        & 0.884 ± 0.000 & \color{blue}+0.0\% 
        & 0.362 ± 0.026 & \color{blue}+5.6\% 
        & 0.439 ± 0.000 & \color{blue}+0.0\% 
        & 0.502 ± 0.006 & \color{blue}+0.0\%
        & 0.622 ± 0.000 & \color{red}-0.7\%
        \\ \midrule
        
        \multirow{2}{*}{\textbf{CMA-ES}} & Base 
        & -0.001 ± 0.014 & 
        & 0.717 ± 0.001 &  
        & 0.389 ± 0.006 & 
        & 0.633 ± 0.015  & 
        & 0.528 ± 0.017 & 
        & 0.633 ± 0.000 & \\
        & \ourmethod 
        & 0.001 ± 0.012 & \color{blue}+0.2\% 
        & 0.717 ± 0.002 & \color{blue}+0.0\% 
        & 0.391 ± 0.006 & \color{blue}+0.2\% 
        & 0.635 ± 0.020 & \color{blue}+0.2\% 
        & 0.527 ± 0.013 & \color{red}-0.1\%
        & 0.633 ± 0.000 & \color{blue}+0.0\%
        \\ \midrule
 
         \multirow{2}{*}{\textbf{GA}} & Base 
         & 0.180 ± 0.019 & 
         & 0.749 ± 0.026 & 
         & 0.474 ± 0.022 & 
         & 0.802 ± 0.022 & 
         & 0.512 ± 0.007 & 
         & 0.633 ± 0.000 & \\
         & \ourmethod 
         & 0.180 ± 0.016 & \color{blue}+0.0\% 
         & 0.755 ± 0.034 & \color{blue}+0.6\% 
         & 0.486 ± 0.019 & \color{blue}+1.2\% 
         & 0.783 ± 0.039 & \color{red}-1.9\% 
         & 0.508 ± 0.004 & \color{red}-0.4\%
         & 0.633 ± 0.000 & \color{blue}+0.0\%
         \\ \midrule
 
        \multirow{2}{*}{\textbf{ENS-MIN}} & Base 
        & 0.220 ± 0.012 & 
        & 0.808 ± 0.013 & 
        & 0.485 ± 0.017 & 
        & 0.821 ± 0.010 & 
        & 0.506 ± 0.009 & 
        & 0.633 ± 0.000 & \\
        & \ourmethod
        & 0.226 ± 0.013 & \color{blue}+0.6\% 
        & 0.819 ± 0.007 & \color{blue}+1.1\% 
        & 0.485 ± 0.011 & \color{blue}+0.0\% 
        & 0.801 ± 0.008 & \color{blue}+2.0\% 
        & 0.507 ± 0.007 & \color{blue}+0.1\%
        & 0.633 ± 0.000 & \color{blue}+0.0\%
        \\ \midrule
 
        \multirow{2}{*}{\textbf{ENS-MEAN}} & Base 
        & 0.220 ± 0.010 & 
        & 0.828 ± 0.019 & 
        & 0.487 ± 0.026 & 
        & 0.806 ± 0.023 & 
        & 0.503 ± 0.007 & 
        & 0.633 ± 0.000 & \\
        & \ourmethod 
        & 0.220 ± 0.008 & \color{blue}+0.0\% 
        & 0.809 ± 0.021 & \color{red}-1.9\% 
        & 0.488 ± 0.009 & \color{blue}+0.1\% 
        & 0.795 ± 0.021 & \color{red}-1.1\% 
        & 0.502 ± 0.007 & \color{red}-0.1\%
        & 0.633 ± 0.000 & \color{blue}+0.0\% \\ \midrule  
 
        \multirow{2}{*}{\textbf{REINFORCE}} & Base 
        & 0.169 ± 0.031 & 
        & 0.479 ± 0.187 & 
        & 0.473 ± 0.015 & 
        & 0.564 ± 0.019 & 
        & 0.512 ± 0.009 & 
        & 0.633 ± 0.000 & \\
        & \ourmethod 
        & 0.174 ± 0.032 & \color{blue}+0.5\% 
        & 0.712 ± 0.008 & \color{blue}+23.3\% 
        & 0.468 ± 0.008 & \color{red}-0.5\%
        & 0.581 ± 0.023 & \color{blue}+1.7\% 
        & 0.511 ± 0.003 & \color{red}-0.1\%
        & 0.633 ± 0.000 & \color{blue}+0.0\%
        \\ \midrule
 
        \multirow{2}{*}{\textbf{MINs}} & Base 
        & 0.738 ± 0.024 & 
        & 0.905 ± 0.003 & 
        & 0.363 ± 0.023 & 
        & 0.511 ± 0.015 & 
        & 0.506 ± 0.007 & 
        & 0.633 ± 0.000 & \\
        & \ourmethod  
        & 0.737 ± 0.014 & \color{red}-0.1\% 
        & 0.903 ± 0.003 & \color{red}-0.2\% 
        & 0.372 ± 0.015 & \color{blue}+0.9\% 
        & 0.519 ± 0.016 & \color{blue}+0.9\% 
        & 0.506 ± 0.006 & \color{blue}+0.0\%
        & 0.633 ± 0.000 & \color{blue}+0.0\%
        \\ \midrule
 
        \multirow{2}{*}{\textbf{COMs}} & Base 
        & 0.624 ± 0.025 & 
        & 0.884 ± 0.004 & 
        & 0.404 ± 0.018 & 
        & 0.714 ± 0.084 & 
        & 0.512 ± 0.013 & 
        & 0.633 ± 0.000 & \\
        & \ourmethod  
        & 0.614 ± 0.031 & \color{red}-1.0\% 
        & 0.881 ± 0.002 & \color{red}-0.3\% 
        & 0.404 ± 0.010 & \color{blue}+0.0\% 
        & 0.746 ± 0.062 & \color{blue}+3.2\% 
        & 0.514 ± 0.011 & \color{blue}+0.2\%
        & 0.633 ± 0.000 & \color{blue}+0.0\%
        \\ \midrule

        \multirow{2}{*}{\textbf{RoMA}} & Base 
         & 0.282 ± 0.025 &
         & 0.724 ± 0.018 &
         & 0.387 ± 0.016 &
         & 0.614 ± 0.068 &
         & 0.525 ± 0.003 &
         & 0.633 ± 0.000 & \\
        & \ourmethod 
        & 0.278 ± 0.018 & \color{red}-0.4\% 
        & 0.725 ± 0.018 & \color{blue}+0.1\% 
        & 0.385 ± 0.018 & \color{red}-0.2\% 
        & 0.647 ± 0.037 & \color{blue}+3.3\% 
        & 0.526 ± 0.003 & \color{blue}+0.1\%
        & 0.633 ± 0.000 & \color{blue}+0.0\%
        \\ \midrule

        \multirow{2}{*}{\textbf{ICT}} & Base 
         & 0.672 ± 0.025 &
         & 0.891 ± 0.003 &
         & 0.405 ± 0.023 &
         & 0.690 ± 0.049 &
         & 0.547 ± 0.016 &
         & 0.633 ± 0.000 & \\
        & \ourmethod
        & 0.685 ± 0.012 & \color{blue}+1.3\% 
        & 0.891 ± 0.004 & \color{blue}+0.0\% 
        & 0.391 ± 0.033 & \color{red}-1.4\% 
        & 0.683 ± 0.035 & \color{red}-0.7\% 
        & 0.555 ± 0.017 & \color{blue}+0.8\%
        & 0.633 ± 0.000 & \color{blue}+0.0\%
        \\ \bottomrule
    \end{tabular}
}
\label{tb:75th_results}
\end{table}

\begin{table}[tbh]
\caption{Percentage of performance improvement achieved by \ourmethod~across all tasks and baselines at the \textbf{$50$-th percentile} level. \textbf{P} denotes the normalized performance while \textbf{G} denotes \ourmethod's percentage gain over the baseline performance.}
    \centering
    \setlength\tabcolsep{3pt} 
    \resizebox{\textwidth}{!}{
        \begin{tabular}{ll|ll|ll|ll|ll|ll|ll} 
         \toprule
           & & \multicolumn{6}{c|}{\textbf{Continuous Tasks}} & \multicolumn{6}{c}{\textbf{Discrete Tasks}} 
           \\\cmidrule(lr){3-8}\cmidrule(lr){9-14}
           & & \multicolumn{2}{c}{\textbf{Ant Morphology}} & \multicolumn{2}{c}{\textbf{D'Kitty Morphology}} & \multicolumn{2}{c|}{\textbf{Superconductor}} 
          & \multicolumn{2}{c}{\textbf{TF Bind 8}} & \multicolumn{2}{c}{\textbf{TF Bind 10}} & \multicolumn{2}{c}{\textbf{ChEMBL}}
          
          \\ \cmidrule(lr){3-4}\cmidrule(lr){5-6} \cmidrule(lr){7-8}\cmidrule(lr){9-10}\cmidrule(lr){11-12}\cmidrule(lr){13-14}

          \multicolumn{2}{c|}{\textbf{Algorithms}}& \multicolumn{1}{c}{\textbf{P}} & \multicolumn{1}{c}{\textbf{G}}
           & \multicolumn{1}{c}{\textbf{P}} & \multicolumn{1}{c}{\textbf{G}}
           & \multicolumn{1}{c}{\textbf{P}} & \multicolumn{1}{c|}{\textbf{G}}
           & \multicolumn{1}{c}{\textbf{P}} & \multicolumn{1}{c}{\textbf{G}}
           & \multicolumn{1}{c}{\textbf{P}} & \multicolumn{1}{c}{\textbf{G}}
           & \multicolumn{1}{c}{\textbf{P}} & \multicolumn{1}{c}{\textbf{G}}
          \\ \midrule 
         \(\mathfrak{D}\)(\textbf{best}) & & 0.565 & & 0.884 & & 0.400 & & 0.439 & & 0.467 & & 0.605
         \\ \midrule
          
        \multirow{2}{*}{\textbf{CbAS}} & Base  
        & 0.371 ± 0.017  & 
        & 0.737 ± 0.021  & 
        & 0.119 ± 0.017  & 
        & 0.426 ± 0.021   & 
        & 0.456 ± 0.006  & 
        & 0.633 ± 0.000  & \\ 
        & \ourmethod   
        & 0.381 ± 0.027 &  \color{blue}+1.0\% 
        & 0.729 ± 0.023& \color{red}-0.8\% 
        & 0.128 ± 0.006& \color{blue}+0.9\% 
        & 0.420 ± 0.026& \color{red}-0.6\% 
        & 0.465 ± 0.009& \color{blue}+0.9\%
        & 0.633 ± 0.000 & \color{blue}+0.0\%
        \\ \midrule 
        
        \multirow{2}{*}{\textbf{BO-qEI}} & Base  
        & 0.568 ± 0.000  & 
        & 0.883 ± 0.000  & 
        & 0.295 ± 0.006  & 
        & 0.439 ± 0.000   & 
        & 0.467 ± 0.000   & 
        & 0.609 ± 0.031 & \\
        & \ourmethod 
        & 0.568 ± 0.000 & \color{blue}+0.0\% 
        & 0.883 ± 0.000 & \color{blue}+0.0\% 
        & 0.305 ± 0.015 & \color{blue}+1.0\% 
        & 0.439 ± 0.000 & \color{blue}+0.0\% 
        & 0.467 ± 0.000 & \color{blue}+0.0\%
        & 0.569 ± 0.000  & \color{red}-4.0\%
        \\ \midrule 
        
        \multirow{2}{*}{\textbf{CMA-ES}} & Base 
        & -0.047 ± 0.005  & 
        & 0.685 ± 0.011  & 
        & 0.378 ± 0.006  & 
        & 0.546 ± 0.011   & 
        & 0.486 ± 0.019  & 
        & 0.633 ± 0.000 & \\
        & \ourmethod  
        & -0.041 ± 0.007 & \color{blue}+0.6\%
        & 0.685 ± 0.010 & \color{blue}+0.0\% 
        & 0.379 ± 0.006 & \color{blue}+0.1\% 
        & 0.550 ± 0.015 & \color{red}-0.4\% 
        & 0.481 ± 0.012 & \color{red}-0.5\%
        & 0.633 ± 0.000 & \color{blue}+0.0\%\\ \midrule 
        
        \multirow{2}{*}{\textbf{GA}} & Base  
        & 0.140 ± 0.018  & 
        & 0.616 ± 0.124  & 
        & 0.459 ± 0.034 & 
        & 0.613 ± 0.034 & 
        & 0.472 ± 0.004 & 
        & 0.633 ± 0.000  & \\
        & \ourmethod
        & 0.141 ± 0.020 & \color{blue}+0.1\% 
        & 0.583 ± 0.158 & \color{red}-3.3\% 
        & 0.471 ± 0.020 & \color{blue}+1.2\% 
        & 0.598 ± 0.043 & \color{red}-1.5\% 
        & 0.468 ± 0.005 & \color{red}-0.4\%
        & 0.633 ± 0.000 & \color{blue}+0.0\%
        \\ \midrule 
        
        \multirow{2}{*}{\textbf{ENS-MIN}} & Base  
        & 0.183 ± 0.009  & 
        & 0.751 ± 0.018  & 
        & 0.478 ± 0.020  & 
        & 0.659 ± 0.023  & 
        & 0.469 ± 0.003  & 
        & 0.633 ± 0.000  & \\
        & \ourmethod  
        & 0.187 ± 0.011 & \color{blue}+0.4\% 
        & 0.758 ± 0.016 & \color{blue}+0.7\% 
        & 0.475 ± 0.013 & \color{red}-0.3\% 
        & 0.624 ± 0.028 & \color{red}-2.5\%
        & 0.469 ± 0.002 & \color{blue}+0.0\%
        & 0.633 ± 0.000 & \color{blue}+0.0\%
        \\ \midrule
        
        \multirow{2}{*}{\textbf{ENS-MEAN}} & Base  
        & 0.182 ± 0.012  & 
        & 0.777 ± 0.028  & 
        & 0.480 ± 0.030  & 
        & 0.632 ± 0.023  & 
        & 0.469 ± 0.002  & 
        & 0.633 ± 0.000  & \\
        & \ourmethod   
        & 0.182 ± 0.004 & \color{blue}+0.0\%
        & 0.751 ± 0.033 & \color{red}-2.6\% 
        & 0.480 ± 0.009 & \color{blue}+0.0\% 
        & 0.629 ± 0.019 & \color{red}-0.3\% 
        & 0.467 ± 0.002 & \color{red}-0.2\%
        & 0.633 ± 0.000 & \color{blue}+0.0\%
        \\ \midrule    
        
        \multirow{2}{*}{\textbf{REINFORCE}} & Base  
        & 0.137 ± 0.022  & 
        & 0.454 ± 0.192  & 
        & 0.468 ± 0.014  & 
        & 0.440 ± 0.010  & 
        & 0.468 ± 0.007  & 
        & 0.633 ± 0.000 & \\
        & \ourmethod  
        & 0.144 ± 0.029 & \color{blue}+0.7\% 
        & 0.697 ± 0.011 & \color{blue}+24.3\% 
        & 0.464 ± 0.008 & \color{red}-0.4\% 
        & 0.455 ± 0.022 & \color{blue}+1.5\% 
        & 0.472 ± 0.005 & \color{blue}+0.4\%
        & 0.633 ± 0.000 & \color{blue}+0.0\%
        \\ \midrule 
        
        \multirow{2}{*}{\textbf{MINs}} & Base  
        & 0.631 ± 0.033  & 
        & 0.887 ± 0.005  & 
        & 0.333 ± 0.019  & 
        & 0.425 ± 0.010  & 
        & 0.468 ± 0.005  & 
        & 0.633 ± 0.000 & \\
        & \ourmethod  
        & 0.628 ± 0.019 & \color{red}-0.3\% 
        & 0.885 ± 0.002 & \color{red}-0.2\% 
        & 0.346 ± 0.013 & \color{blue}+1.3\% 
        & 0.424 ± 0.010 & \color{red}-0.1\% 
        & 0.467 ± 0.006 & \color{red}-0.1\%
        & 0.633 ± 0.000 & \color{blue}+0.0\%
        \\ \midrule 
        
        \multirow{2}{*}{\textbf{COMs}} & Base  
        & 0.502 ± 0.027  & 
        & 0.862 ± 0.005  & 
        & 0.384 ± 0.020  & 
        & 0.593 ± 0.052  & 
        & 0.474 ± 0.010  & 
        & 0.633 ± 0.000 & \\
        & \ourmethod   
        & 0.489 ± 0.026 & \color{red}-1.3\% 
        & 0.854 ± 0.003 & \color{red}-0.8\% 
        & 0.383 ± 0.013 & \color{red}-0.1\%
        & 0.634 ± 0.061 & \color{blue}+4.1\% 
        & 0.476 ± 0.012 & \color{blue}+0.2\%
        & 0.633 ± 0.000 & \color{blue}+0.0\%
        \\ \midrule

        \multirow{2}{*}{\textbf{RoMA}} & Base 
         & 0.227 ± 0.013 &
         & 0.612 ± 0.146 &
         & 0.356 ± 0.024 &
         & 0.489 ± 0.060 &
         & 0.518 ± 0.005 &
         & 0.633 ± 0.000 & \\
        & \ourmethod 
        & 0.224 ± 0.013 & \color{red}-0.3\% 
        & 0.636 ± 0.089 & \color{blue}+2.4\% 
        & 0.358 ± 0.021 & \color{blue}+0.2\% 
        & 0.512 ± 0.051 & \color{blue}+2.3\% 
        & 0.516 ± 0.006 & \color{red}-0.2\%
        & 0.633 ± 0.000 & \color{blue}+0.0\%
        \\ \midrule

        \multirow{2}{*}{\textbf{ICT}} & Base 
         & 0.547 ± 0.024 &
         & 0.870 ± 0.004 &
         & 0.374 ± 0.021 &
         & 0.590 ± 0.038 &
         & 0.505 ± 0.014 &
         & 0.633 ± 0.000 & \\
        & \ourmethod
        & 0.560 ± 0.018 & \color{blue}+1.3\% 
        & 0.870 ± 0.005 & \color{blue}+0.0\% 
        & 0.363 ± 0.033 & \color{red}-1.1\% 
        & 0.575 ± 0.030 & \color{red}-1.5\% 
        & 0.518 ± 0.013 & \color{blue}+1.3\%
        & 0.633 ± 0.000 & \color{blue}+0.0\%
 \\ \bottomrule
    \end{tabular}
}
\label{tb:50th_results}
\end{table}

\subsection{Setting value for $\lambda$:}
\label{app:tuning_lambda}
\begin{table}[ht]
    \begin{minipage}{.4\textwidth}
        \centering
        \begin{tabular}{c|c|c}
        \toprule
        \vspace{0.4cm}
        $\lambda$ & \textbf{Ant-Morphology} & \textbf{TF Bind 8}     \\ \midrule
        1e-4      & 0.295 ± 0.026  & 0.976 ± 0.014 \\
        1e-3      & \textbf{0.314 ± 0.034}  & \textbf{0.986 ± 0.007} \\
        1e-2      & 0.307 ± 0.026  & 0.967 ± 0.015 \\
        1e-1      & 0.299 ± 0.038  & 0.968 ± 0.017 \\ \bottomrule
        \end{tabular}
        \caption{Hyper-parameter tuning for $\lambda$ when \ourmethod~ is applied to the \textbf{GA} baseline on two tasks: \textbf{Ant-Morphology} and \textbf{TF Bind 8}.}
        \label{tb:tune_lambda}
    \end{minipage}%
    \hfill
    \begin{minipage}{.57\textwidth}
      \centering
        \begin{tabular}{c|c|c}
        \toprule
        Bound $\omega_\mu$ & Bound $\omega_\sigma$ & Optimization performance \\
        \midrule
        $[-10^-3,10^-3]$   & $[10^-5,10^-2]$       & \textbf{0.656 ± 0.042}            \\
        $[-0.1,0.1]$       & $[10^-3,0.1]$         & 0.626 ± 0.035            \\
        $[-0.5,0.5]$       & $[10^-3,0.5]$         & 0.636 ± 0.043            \\
        $[-1,1]$           & $[10^-3,1]$           & 0.636 ± 0.028            \\
        $[-2,2]$           & $[10^-3,2]$           & 0.632 ± 0.036            \\
        $[-4,4]$           & $[10^-3,4]$           & 0.620 ± 0.028           \\ \bottomrule
        \end{tabular}
        \caption{Experiments on \textbf{TF-BIND-10} using \textbf{Cbas} regularized with our BOSS regularizer, varying the bounds/limits for $\omega_\mu$ and $\omega_\sigma$.}
        \label{tb:large_omega_effect}
    \end{minipage}
\end{table}


The regularization coefficient $\lambda$ controls the balance between the original loss and the proposed regularizer. It must be set carefully to avoid being too small or too large, ensuring an effective trade-off between the two terms.

In practice, we find $\lambda = 10^{-3}$ to be reasonably effective across all task benchmarks. We conduct hyper-parameter tuning for $\lambda$ when \ourmethod~ is applied to the \textbf{GA} baseline on two tasks: \textbf{Ant-Morphology} and \textbf{TF Bind 8}. The results are reported in Table \ref{tb:tune_lambda}.

\subsection{Convergence Analysis}
\label{app:convergence_analysis}
Our perturbation distribution is learned to optimize our notion of sensitivity while simultaneously minimizing the fit of the surrogate. This results in a min-max optimization task of a non-convex function $\min_\phi \max_\omega F(\phi, \omega)$ which is then optimized via stochastic gradient descent ascent (SGDA) for $(\phi, \omega)$ -- see Algorithm \ref{alg:pseudo-code}. Although SGDA can generally have convergence issues, this is not the case here due to the specific form of our loss function, $F(\phi, \omega) = L(\phi) + \lambda \cdot S_\phi(\omega)$

To elaborate, suppose the gradient norm of the sensitivity measure $\|\nabla_\omega S_\phi(\omega)\|^2$ is bounded below by $2\mu$ with $\mu>0$, then
\begin{eqnarray}
    \frac{1}{2} \| \nabla_\omega F(\phi, \omega) \|^2 &=& \frac{1}{2} \lambda^2 \| \nabla \omega S_\phi(\omega)\|^2 \nonumber \\
    &\geq& \lambda^2 \mu \geq \lambda^2 \mu (S_\phi(\omega^\ast) - S_\phi(\omega)) \nonumber \\
    &=& \lambda \mu (F(\phi, \omega^\ast) - F(\phi, \omega))
\end{eqnarray}
where $\omega^\ast \triangleq \arg\max S_\phi(\omega)$ and the second last step holds because $S_\phi$ is defined to be a probability mass in $(0, 1)$ -- see Definition \ref{def:1} -- and hence, $1 \geq (S_\phi(\omega^\ast) - S_\phi(\omega)) \geq 0$. This final equation is known as the Polyak-Łojasiewicz (PL) condition with constant $\lambda\mu$. Along with the commonly adopted assumptions on the smoothness and Lipschitz continuity of $F(\phi, \omega)$, it implies that SGDA will converge, following Theorem 6.1 of \citep{deng2021local}.

It's important to note that Theorem 6.1 of \citep{deng2021local} is originally positioned in the distributed optimization context, which reduces to the centralized case when the number of optimizing clients is set to 1. Additionally, the theorem references prior work that directly proves the convergence of vanilla local SGDA on non-convex min-max optimization tasks with the PL condition.

This explains why our alternating optimization scheme (i.e., local SGDA) will converge under mild technical assumptions (i.e., a lower-bounded gradient of sensitivity, and smooth loss function with Lipschitz continuity). These assumptions are reasonable and have been introduced in prior work on convergence analysis.

\subsection{Effectiveness of Eq. (12)}
Assuming $\Phi$ accurately approximates the sensitivity $S_\phi$ via Eq. (8)-(9), Eq. (12) is a mathematically correct application of the chain rule. Therefore, the effectiveness of Eq. (12) is tied to the approximation accuracy of $S_\phi$ via Eq. (9).  This is explained as follows:

According to Eq. (9), the sensitivity measure (see Eq. (4)) is defined as the expectation of $\Phi(\gamma)$, where the expectation is taken over the distribution of $\gamma$. This expectation is approximated using an empirical average based on a finite set of $m$ samples of $\gamma$.

The learning accuracy of the sensitivity measure can be characterized by the gap between the expectation and the empirical average of a random variable $\mathbf{I}(\gamma \in \mathfrak{R}_\alpha(\phi)) \in [0, 1]$ based on a finite sample set. This gap can be upper-bounded using the Hoeffding bound \citep{hoeffding1994probability}, which implies a learning accuracy of $\epsilon$ with $m = O(1/\epsilon)$ samples of $\gamma$. In practice, we find that $m = 100$ samples are sufficient.

\subsection{Effect if the upper limit of $\omega$ is too large}
\label{app:large_omega_bound}
To investigate this effect, we conducted multiple experiments on \textbf{TF-BIND-10} using \textbf{Cbas} regularized with our BOSS regularizer, varying the bounds/limits for $\omega_\mu$ and $\omega_\sigma$. The results are reported in Table \ref{tb:large_omega_effect}.

The first row in the Table \ref{tb:large_omega_effect} uses the same bound setting as in our main paper, which appears to yield the best performance. Overall, the results suggest that increasing the range of the bounds can cause the sampled perturbations to have larger values. These larger perturbations might dominate the learning of the surrogate parameters, meaning the direction of the gradient update is primarily influenced by $\omega$, leading to minimal changes in the gradient direction for the $\phi$ component. This slows down the learning of $\phi$ and ultimately reduces the final optimization performance. Note that we have previously shown that the predictive accuracy of $\phi$ correlates with its optimization performance.

\subsection{Connection to robust optimization}
There is a conceptual connection between our work and robust optimization/machine learning through the idea of sensitivity measures, though these ideas are substantiated in different mathematical forms tailored to different use cases. This is elaborated below.

First, our work relates to robust optimization and adversarial learning through the high-level concept of sensitivity measures. However, the mathematical formulations differ to suit different applications. Intuitively, sensitivity can be viewed as brittleness to low-energy adversarial noise, a well-known concept in adversarial or robust machine learning. In our definition of sensitivity, the perturbation $\gamma$ corresponds to adversarial noise in adversarial learning. The distinction lies in the formulation tailored to different use cases: In adversarial machine learning, sensitivity/brittleness is defined based on the test input and is used post-training to attack a previously trained model on a specific test input. In our problem setting, the definition applies to the model weights and also depends on the entire training input distribution, used during training to condition the surrogate model.

Second, beyond its main use of conditioning the surrogate model to improve the performance of existing offline optimizers, our sensitivity measurement (see Eq. (4)) could potentially serve as a regularizer in training modern deep learning models. This might enable a form of certified (attack-agnostic) defense in scenarios where the test distribution is known in advance. Most defense approaches in robust machine learning assume knowledge of the test distribution. For instance, if we replace the training distribution with the test distribution over which the expectation is taken in Eq. (4), the sensitivity measurement could be seen as a certificate condition: For any Gaussian noise from a given distribution, the expected output of the perturbed model will not significantly differ from the expected output of the unperturbed model with high probability. This is a speculative hypothesis about a potential connection or direction. Whether this notion of a robust certificate would be accepted by experts in the field of robust machine learning requires further investigation, which is beyond the scope of this paper.

\subsection{More RMSE curves in training and OOD regimes}

\begin{figure}[t]
     \centering
     \begin{minipage}{0.48\linewidth} 
        \centering
         \includegraphics[width=\textwidth]{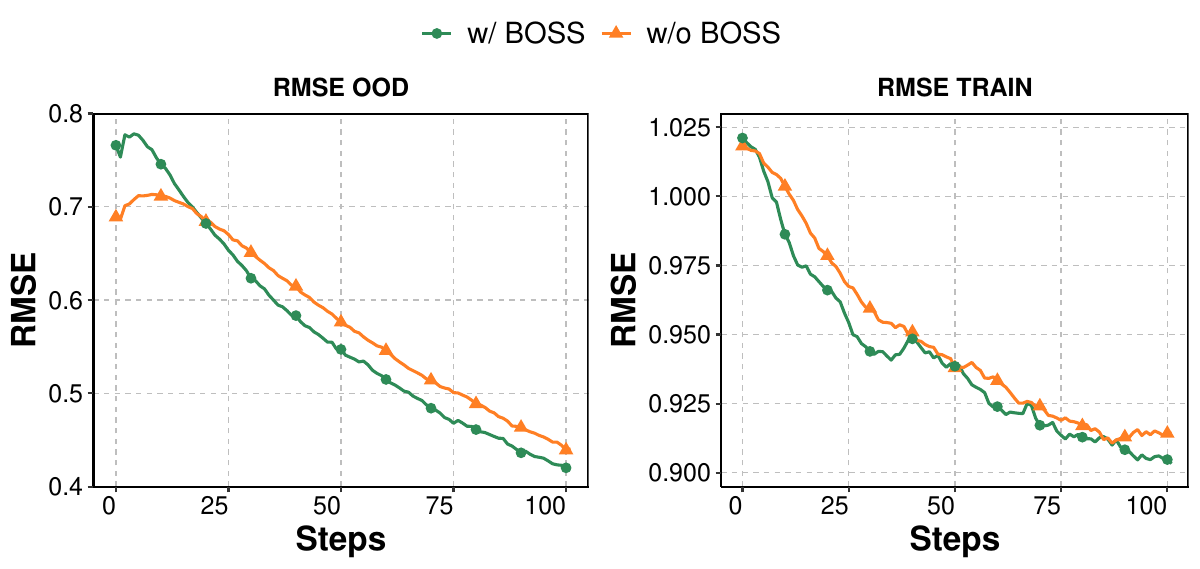}
         (a)
         \label{fig:COMs_tf8}
     \end{minipage}
     \hfill
     \begin{minipage}{0.48\linewidth} 
        \centering
         \includegraphics[width=\textwidth]{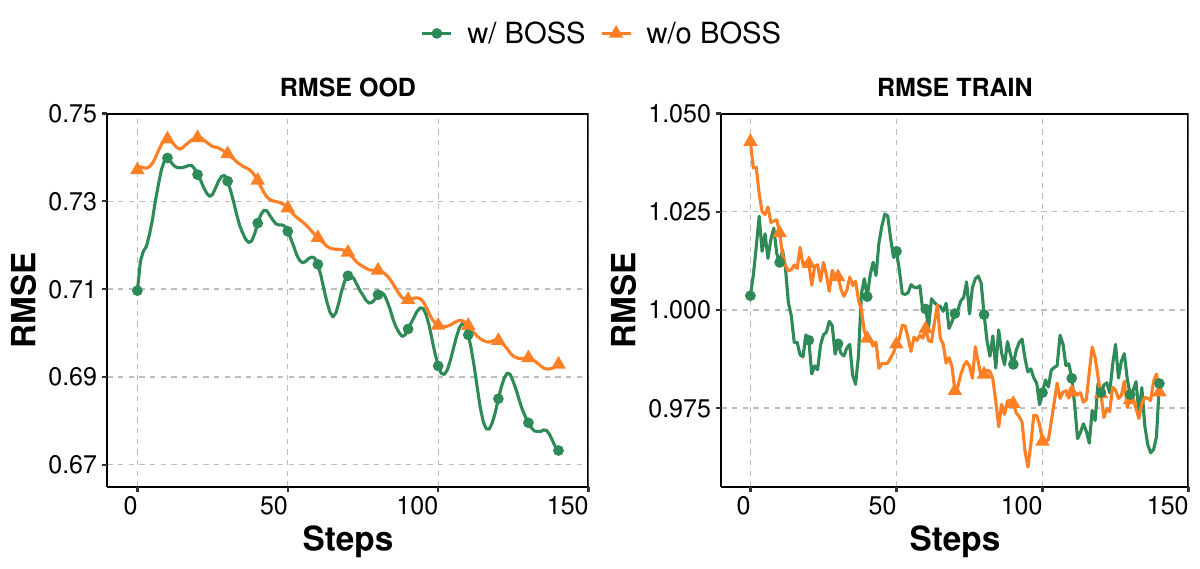}
         (b)
        \label{fig:COMs_tf10}
     \end{minipage}
     
     \caption{RMSE curves of \textbf{COMs}'s surrogate model, which is trained on the offline data of \textbf{TF-BIND-8} (a) and \textbf{TF-BIND-10} (b), on both training and unseen test data. RMSE OOD denotes the RMSE curves on unseen data while RMSE TRAIN denotes the RMSE curves on training data}
     \label{fig:more_rmse_curve}
\end{figure}

Similar to Figure \ref{fig:additional_exp}b which shows the RMSE curves for \textbf{CbAS}'s surrogate model trained on the offline data of \textbf{TF-BIND-8} and \textbf{TF-BIND-10}, for both training and unseen test data, we also plot the RMSE curves for \textbf{COMs}'s surrogate on the same datasets. The plots are illustrated in Figure \ref{fig:more_rmse_curve}. The results clearly demonstrate that our regularizer, \ourmethod, effectively reduces RMSE on both training and test data, thereby improving overall performance.

\subsection{Limitations}
\label{app:limitation}
We want to clarify that the consistently improved performance of our method is not unexpected. Unlike previous work that proposes a competing method and may struggle to outperform all existing methods across all tasks, our approach acts as a booster, enhancing the performance of existing methods.

This aspect is both a strength and a limitation. It is a strength because it can robustly and directly leverage prior work to advance the state-of-the-art and is arguably applicable as a model-agnostic booster to future methods. However, it is also a limitation because it is not a standalone method; its effectiveness depends on the base performance of the existing method it is coupled with. Additionally, it is only compatible with methods characterized by a differentiable loss function, and cannot be applied to methods involving non-differentiable training routines.

Moreover, the booster’s effectiveness diminishes when considering lower solution percentiles, such as the $50$-th and $75$-th percentiles, as detailed in Tables \ref{tb:50th_results} and \ref{tb:75th_results}. The chance of a successful boost in these cases is substantially lower than that for the $100$-th percentile, as reported in Table \ref{tb:100th_results}.

\end{document}